\documentclass[a4paper,onesided,12pt]{report}
\usepackage{styles/fbe_tez}
\usepackage[utf8x]{inputenc} 

\usepackage{amsmath, amsthm, amssymb}
\usepackage[bottom]{footmisc}
\usepackage{cite}
\usepackage{graphicx}
\usepackage{longtable}
\graphicspath{{figures/}} 
\usepackage{breakurl}
\usepackage[breaklinks]{hyperref}
\usepackage{multirow}

\usepackage{subfigure}
\usepackage{algorithm}
\usepackage{algorithmic}
\usepackage{adjustbox}
\usepackage{float}
\usepackage{courier}
\usepackage{hyperref}
\usepackage{graphics}
\usepackage{color, colortbl}
\usepackage{textcomp}
\definecolor{Gray}{gray}{0.9}
\definecolor{LightCyan}{rgb}{0.88,1,1}
\usepackage[first=0,last=9]{lcg}

\title{ANALYZING THE GENERALIZABILITY OF DEEP CONTEXTUALIZED LANGUAGE REPRESENTATIONS FOR TEXT CLASSIFICATION
}
\turkcebaslik{DOĞAL DİL İŞLEMEDE BAĞLAM ODAKLI DERİN ÖĞRENME YÖNTEMLERİNİN TRANSFER KAPASİTESİNİN METİN SINIFLANDIRMASI PROBLEMİ ÜZERİNDEN İNCELENMESİ}
\degree{B.S., Computer Engineering, Boğaziçi University, 2016}
\author{Berfu Büyüköz}
\program{Computer Engineering}
\subyear{2020}

\supervisor{Assoc. Prof. Arzucan Özgür Türkmen}
\cosuperi{Ali Hürriyetoğlu, Ph.D.}
\examineri{Prof. Tunga Güngör}
\examinerii{Assist. Prof. Emre Uğur}
\examineriii{Prof. Deniz Yüret}
\dateofapproval{}

\begin{document}

\pagenumbering{roman}
\makemstitle 
\makeapprovalpage
\begin{acknowledgements}
I am grateful to my family for their unconditional love and patience. I am grateful to Arzucan Özgür, for being such an inspiring figure by her selfless devotion to research the most righteous way with the passion to contribute to the community. I am grateful to Ali Hürriyetoğlu, for being such a role model, who could somehow always find a way to turn the mist of research questions into a structured path to create practical solutions by combining creativity and technique. I cannot thank enough my dear friends who put up with my whims throughout this journey. I thank fellows from TabiLAB for inspiring me with their brilliance, invaluable insights and recommendations. I thank Koç University EMW research team for their generosity in sharing the data which was created with blood, sweat and tears. I feel lucky that I got to meet fellows in EMW project engineering team who invested their precious time and energy to support me in this study from the very beginning. Lastly, I owe the deepest gratitude to our professors and staff members in our department who taught us how to form such a great community and made it feel like the dearest home from the day one.

The numerical calculations reported in this paper were partially performed at TUBITAK ULAKBIM, High Performance and Grid Computing Center (TRUBA resources).
\end{acknowledgements}

\begin{abstract}
This study evaluates the robustness of two state-of-the-art deep contextual language representations, ELMo and DistilBERT, on supervised learning of binary protest news classification and sentiment analysis of product reviews. A ``cross-context'' setting is enabled using test sets that are distinct from the training data. Specifically, in the news classification task, the models are developed on local news from India and tested on the local news from China. In the sentiment analysis task, the models are trained on movie reviews and tested on customer reviews. This comparison is aimed at exploring the limits of the representative power of today’s Natural Language Processing systems on the path to the systems that are generalizable to real-life scenarios. The models are fine-tuned and fed into a Feed-Forward Neural Network and a Bidirectional Long Short Term Memory network. Multinomial Naive Bayes and Linear Support Vector Machine are used as traditional baselines. The results show that, in binary text classification, DistilBERT is significantly better than ELMo on generalizing to the cross-context setting. ELMo is observed to be significantly more robust to the cross-context test data than both baselines. On the other hand, the baselines performed comparably well to ELMo when the training and test data are subsets of the same corpus (no cross-context). DistilBERT is also found to be 30\% smaller and 83\% faster than ELMo. The results suggest that DistilBERT can transfer generic semantic knowledge to other domains better than ELMo. DistilBERT is also favorable in incorporating into real-life systems for it requires a smaller computational training budget. When generalization is not the utmost preference and test domain is similar to the training domain, the traditional ML algorithms can still be considered as more economic alternatives to deep language representations.   
\end{abstract}

\begin{ozet}
Bu çalışma ELMo ve DistilBERT adındaki bağlamsal ve derin doğal dil sistemlerini protesto haber metni ve kullanıcı yorumlarının ikili sınıflandırılması olmak üzere iki farklı senaryo üzerinden kıyaslamaktadır. Asıl amaç, bu modern sistemlerin birbirinden çok farklı girdileri modellemedeki başarısını ölçmek, bu sayede doğal dil işlemeyi hedef alan sistemlerin gerçek hayattaki kaynak çeşitliliğine ne kadar iyi adapte olabildiğine ışık tutmaktır. Bu amaçla, modeller eğitilirken ve başarımı ölçülürken bağlamca ayrışan veri kümeleri kullanılmıştır. Sözgelimi, ilk senaryo için Hindistan ve Çin'in yerel gazete haberlerinden, ikinci senaryo için ise sinema filmleri ve teknolojik cihazlara yapılan kullanıcı yorumlarından faydalanılmıştır. ELMo ve DistilBERT kullanılarak üretilen kelime vektörleri, biri ileriye doğru, diğeri özyineli olmak üzere iki farklı sinir ağına verilmiştir. Daha basit ve bağlamsal olmayan Çokterimli Naif Bayes Sınıflandırıcısı ve Doğrusal Destek Vektör Makinesinin sonuçları temel alınmıştır. Sonuçta, DistilBERT'in ELMo'ya kıyasla, bu iki senaryoda, farklı test ortamlarına daha dayanıklı olduğu, üstüne \%30 daha küçük ve \%83 daha hızlı olduğu görülmüştür. ELMo ise, yine, farklı bağlama geçişte, iki temel algoritmadan daha başarılı olmuştur. Buradan yola çıkarak, DistilBERT'in dilin genellenebilir anlamsal özelliklerini daha iyi öğrendiği ve gerçek-zamanlı dönüt gerektiren sistemlerde ve bellek kaynağı bakımından sınırlı bütçelerde ELMo'dan daha kullanışlı olduğu yargısına varılabilir. Öte yandan, genellenebilirliğin gözetilmediği, test verisinin eğitim verisine benzediği durumlarda Naif Bayes gibi daha basit ve ekonomik algoritmalar derin sinir ağlarına alternatifler olarak tercih edilebilir.
\end{ozet}
\tableofcontents
\listoffigures
\listoftables





\chapter{INTRODUCTION}
\label{chapter:introduction}
\pagenumbering{arabic}

Natural Language Processing (NLP) has been a promising research area for decades, and it has been living its golden years so far with the rise of Neural Networks in Machine Learning (ML).  With the world becoming interconnected more and more, gathering data and setting up the environment has never been that easy. And many different approaches to understand natural languages are proposed and proved effective.

Natural Language Processing is a research area to build computational systems that can interpret the meaning of natural language input in various forms. Understanding a language input, especially if a human is a native speaker, is a trivial process for humans, which continuously takes place on the background of everyday life. But teaching machines to do so is equally hard since this requires humans to break the very process into pieces, and how language acquisition is realized is still a debated question \cite{10.3389/fpsyg.2017.01918}.

As the ultimate task is to build a reasonably well language model \cite{ettinger-generalizability-st}, researchers treated this ultimate goal by dividing it into tasks and attacking each of them separately. There are various NLP tasks, that can be grouped under two main categories: syntactic (e.g. part of speech tagging) and semantic (e.g. sentiment analysis) tasks. These categories are not mutually exclusive that many tasks fall under both, for the meaning is something revealed by structure and semantics combined. Studying the NLP tasks separately comes handy in which it makes it possible to treat the language understanding problem on different levels of input: token-level (e.g. named entity recognition), sentence-level (e.g. textual entailment), and document-level (e.g. anaphora resolution).

A challenge the NLP community faces today is to leverage NLP systems from a well-maintained test environment to more realistic scenarios with full of dynamism and diversity \cite{ettinger-generalizability-st, hurriyetoglu-CLEF}. Accuracy and speed are the first criteria that might be naturally expected from a real-life system. But at the same time, an NLP system should generalize well to data coming from diverse sources differing in time and space.

One natural way of imposing robustness to diversity is to create systems using large and diverse data. But this brings yet another challenge, that is, collecting a big task-specific corpus. Even a sufficiently large corpus is available, algorithms may not fully distill the information in the data. Also, even if the gold standard, an offline data set could fail to provide time-dynamic information.

Supervised learning schemes using Neural Networks have long been appreciated as an effective way of modeling data across many fields of study, as they eliminate feature engineering and provide a solid way of tackling various NLP tasks. But the success of Neural Network models is still bound to the quality of labeled task-specific training data.

Creating high-quality labeled data is a difficult job that requires domain-specific expert knowledge and lots of time is spent on the quality assessment before launching the data for use. This makes annotation a painful and time-consuming process. Getting through the same process for every other NLP task does not scale in long term solutions. 

The quest of building generalizable systems that are robust and scalable to real-life scenarios in an effective way made the NLP community investigate if they can build task-agnostic models in an unsupervised manner to represent generic syntactic and semantic knowledge of a language. One of the solutions is to create one big universal language representation and use it as the initialization point for any NLP task.

The data preparation bottleneck made the NLP community seek to utilize unlabeled data more. Semi-supervised and self-supervised learning are the products of this quest aimed at incorporating unsupervised language representation techniques more effectively into supervised tasks. 


One example of unsupervised language representations is the famous \texttt{word2vec} \cite{word2vec-NIPS2013_5021}, which creates continuous word vectors for each word in the vocabulary derived from a large corpus in a fully unsupervised manner by utilizing context information regarding the neighboring words. \texttt{word2vec} creates fixed vectors for each unique word in the vocabulary. In this sense, it lacks representing the dynamism of the word meaning that changes depending on the enclosing context. 

The contextualization notion which was yet unable to fully flourish would be the key to create universal language representations that can handle rich syntactic and semantic space of real-life language usage. In this respect, in the last couple of years, several deep contextual neural architectures are proposed and proved surprisingly well on a diverse range of downstream NLP tasks \cite{peters-etal-2018-deep, devlin-etal-2019-bert, radford2019language}.


These new-generation models (e.g. ELMo \cite{peters-etal-2018-deep}, DistilBERT \cite{sanh2019distilbert}) attack the generalization problem by building task-agnostic universal language representations in an unsupervised manner to represent generic syntactic and semantic knowledge of a language, to be then fine-tuned for any NLP task. Recent studies validated that with little fine-tuning data, these representations get results competing with supervised task-specific models on many domains \cite{peters-etal-2018-deep, devlin-etal-2019-bert}.

Contextual representations are successful on a diverse set of NLP tasks, suggesting that they capture generally useful and divergent linguistic information. There is still much to do to understand the true capacity of these representations. Exploring the true limits of what these networks offer is the key to understand how to build next-generation systems and much excitingly, digging into these models might shed light on general language understanding phenomena itself on a cognitive level \cite{grandstrand:2004, KELL2018630}. For this reason, exhaustive evaluation and interpretation studies are needed to be performed on as many different data and task sets as possible.


This study is conducted to contribute to the extrinsic evaluation of the robustness of these representations by testing them on a cross-context data, where the source and target data differ in the originated country and domain \cite{hurriyetoglu-CLEF}. 

More specifically, this study compares internal vector representations of ELMo and DistilBERT on binary protest news classification (PC) and sentiment analysis (SA) of product reviews. 

A ``cross-context'' setting is enabled using test sets that are distinct from the training data. Specifically, in the news classification task, the models are developed on local news from India and tested on the local news from China. In SA, the models are trained on movie reviews and tested on customer reviews. A simpler setting will be referred to as ``null context'' where the models are tested on a subset of the corpora from which the training data sets originate (India and movie reviews test sets).

Using data from different countries and domains makes it possible to create a cross-context setting between training and test phase, which is to use test data that is distinct from both fine-tuning and pretraining data.



This study methodologically deviates from the previous work on assessing the generalizability by using a new socio-political and local news data set other than heavily used data sets. Second, the evaluation is done on a cross-context data (China and customer reviews test sets), without any domain adaptation.  

This study asks the following questions:
\begin{enumerate}
    \item How robust are ELMo and DistilBERT in the cross-context text classification?
    \item Are contextual representations better in the cross-context than much smaller and faster traditional baselines?
    \item Which one is more scalable in terms of model size and training time: ELMo or DistilBERT?
\end{enumerate}

This study draws the following conclusions in binary text classification:
\begin{enumerate}
    \item DistilBERT is more robust than ELMo in the cross-context.
    \item Both ELMo and DistilBERT outperform the baselines (Multinomial Naive Bayes and Linear Support Vector Machine) in generalizing to the cross-context.
    \item DistilBERT is more efficient than ELMo with 30\% smaller size and on average of the two tasks, 83\% faster training and testing time.
    \item Traditional methods like MNB and LSVM can still compete with contextual embeddings when training and test data do not differ very much.
\end{enumerate}

(Note that the conclusions of this study must be considered under the limitations of the experimental setup - See Section \ref{chapter:experimental-setup} for detailed information.)

As the remaining parts of this study, Section \ref{chapter:related-work} outlines the related work, Section \ref{chapter:tasks} and \ref{chapter:data} introduce the tasks and the data sets, Section \ref{chapter:experimental-setup} describes the experimental setup, Section \ref{chapter:experiment-results} reports the experiment results, Section \ref{chapter:discussion} discusses the results, Section \ref{chapter:conclusion} retrospects on what is learnt, and finally Section \ref{chapter:future-work} discourses on the possible improvements and the future work. 
\chapter{RELATED WORK}
\label{chapter:related-work}

This section gives general information about the strengths and weaknesses of various word representations and language representation techniques in the order of their proposition time. Then the previous work that is focusing on understanding the generalization capacity of the contextual language representations is overviewed.

\section{Word vectors in transfer learning scheme}
Inspired by the success of transfer learning using ImageNet in computer vision, research seeking to build generalizable ML systems has gained attention in recent years of the NLP field. Collobert et al. \cite{Collobert:2011:NLP:1953048.2078186} built a generic model trained using a large amount of unlabeled text data for language modeling task to capture internal representations to be then used for various supervised NLP tasks to reduce reliance on apriori task-specific NLP knowledge.

Then emerged two popular procedures, named  Continuous Bag of Words (CBOW) and Skip-Gram models, both to create a language model that computes distributed and continuous word representations from the Google News 6B data set \cite{Mikolov-skip-gram-41224}. The former predicts current word based on the context without utilizing word order information, the latter predicts surrounding words both from history and the future based on the current word by maximizing the average log probability of occurrence of context words. Unlike standard Bag of Words, CBOW uses continuous word vectors. These models first learn continuous word vectors in a simple model in an unsupervised manner and then use them to train an N-gram language model. These new log-linear models can capture subtle sense relations between words that previous Feed-Forward Neural Network (FFNN) and Recurrent Neural Network (RNN) based models miss out as they create high dimensional word vectors by utilizing larger training data sets with much lower computational complexity.

As an example of what CBOW and Skip-gram are capable of, consider the word analogy task of finding the word similar to “Germany” in the same sense “sushi” is similar to “Japan”. To achieve this task, first, a vector representation is computed by the operations V(sushi)-V(Japan)+V(Germany), then by using cosine similarity, the word with the vector closest to the calculated vector is returned as the answer, which becomes surprisingly, the word ``bratwurst''.

It is shown that both Skip-gram and CBOW outperform other previous word representations on both semantic and syntactic relationship tasks, where Skip-gram being better than CBOW in semantic relationship tasks. CBOW is trained much faster and can learn frequent words better but it tends to neglect rare words when predicting the word form context, which is a problem handled more chiefly by design in Skip-Gram, which takes word and context pairs as output candidates without causing a race between rare and frequent words \cite{Mikolov-skip-gram-41224}.

The revolutionary \texttt{word2vec} \cite{word2vec-NIPS2013_5021} is developed as an extension to Skip-gram vectors to create higher quality word vectors by subsampling of the frequent words to alleviate imbalance between rare and frequent words, at the same time speeding up training between 2X and 10X with the help of negative sampling which is the simplified variant of Noise Contrastive Estimation as an alternative to hierarchical softmax. With \texttt{word2vec} creating phrase representations is also made possible, enabling additive compositionality that uncover non-obvious relations such as $V($Russia$) + V($river$) = V($Volga river$)$. It is also illustrated that \texttt{word2vec} finds more fine-grained relationships between words and phrases than prior word representations \cite{Collobert:2011:NLP:1953048.2078186}, also surpassing their performance with a huge margin, requiring a much smaller (a month versus a day) training time \cite{word2vec-NIPS2013_5021}.

GloVe \cite{pennington-etal-2014-glove}, standing for ``global vectors for word representation'', won its well-deserved place as a word representation as effective as \texttt{word2vec} yet with much more simpler and faster training procedure that is realized on non-zero entries of a global word co-occurrence matrix with the training objective of learning word vectors so that their dot products are equal to the logarithm of their co-occurrence. In this sense, GloVe attempts, like \texttt{word2vec} does, to capture nuance relationships that cannot simply be represented by a single scalar such as cosine similarity distance between a pair of vectors. Without using word order and with a much faster training schedule, GloVe achieves to attain matching and exceeding scores of Skip-gram and CBOW algorithms on word analogy, word similarity and named entity recognition tasks.

Skip-gram model is adapted to sentence level by Skip Thought vectors \cite{kiros-skipthought-NIPS2015_5950}, which is an encoder model trained on learning generic sentence representations. This encoder then can be frozen and used off-the-shelf in a diverse set of NLP tasks including semantic-relatedness, paraphrase detection, image-sentence ranking. Skip thought vectors are an attempt to create robust representations in an unsupervised manner to give freedom in the face of the bottleneck of task-specific supervised training.

Last but not least, Bojanowski et al. \cite{bojanowski-etal-2017-enriching} use character N-grams to create word vectors to handle unknown and rare words and to also incorporate morphological information to the word representations.

\section{Contextualized language representations}
One of the first attempts to contextualize pretrained word vectors is CoVe \cite{cove-NIPS2017_7209}, standing for ``context vectors'', to create shared representations for NLP models with the help of an encoder which is a 2-layer bidirectional Long Short Term Memory network (BiLSTM) pretrained on Machine Translation (MT) task. CoVe is inspired by how transfer learning is applied in computer vision by incorporating an ImageNet model to any other computer vision tasks and mimics this process by replacing ImageNet-Convolutional Neural Network pair with MT-LSTM pair.

The prodigy transformer model \cite{transformer-NIPS2017_7181}, solely using attention mechanism, succeeded in outrunning prior sequence transduction models on machine translation tasks which were also relying on convolutions and recurrence relations. The transformer reduces training time by allowing parallelization of computations and generalizes well to other tasks such as constituency parsing with limited training data, and also produces more interpretable models.

Focusing on enhancing universality on the text classification task, text classification fine-tuning scheme called ULMFiT \cite{howard-ruder-2018-universal}, short for ``universal language model fine-tuning'', was proposed that consists of unsupervised pretraining of a generic LSTM-based language model on a large unlabelled data and then transferring knowledge learned by that generic model by supervised fine-tuning on the text classification task.

Radford et al. \cite{radford2019language} pinpoint that LSTM restricts the prediction ability to short-range, thus adapts a transformer-based model to the scheme proposed by \cite{howard-ruder-2018-universal}, and evaluates it on a wider range of tasks including natural language inference, paraphrase detection, and story completion.

It is further shown by Radford et al. \cite{radford2019language} that a language model that is trained on the large WikiText data set \cite{DBLP:conf/iclr/MerityX0S17} performs well across many diverse sets of tasks without the need for explicit supervision. Their biggest transformer model named GPT-2 achieved state-of-the-art results on 7 out of 8 language modeling data sets in a zero-shot setting.

ELMo \cite{peters-etal-2018-deep}, short for ``Embeddings from Language Models'', is one of the most effective contextual word representation models, which is a feature-based word representation aimed at capturing word polysemy within context along with syntactic and semantic characteristics by averaging internal states of a deep bidirectional language model (BiLM) with learned weights (scalar-mixing). 

Then BERT \cite{devlin-etal-2019-bert}, standing for ``Bidirectional Encoder Representations from
Transformers'', emerges with an extraordinary success across many token and sentence-level tasks in the traditional NLP pipeline including part-of-speech tagging, co-reference resolution, dependency labeling, combined with small multi-layer perceptron (MLP) models and without substantial task-specific architectural modifications. BERT owes its success to incorporating bidirectionality to the transformer model trained on large unlabelled data with the masked language modeling (MLM) objective. Unlike \texttt{word2vec}, it learns a context-sensitive representation of a word’s instance within a sentence, by taking the weighted combination of single attention heads, word order, and word position information.

The bidirectionality is made possible by using MLM as the pretraining objective, in which random words from both left and right context are masked and the model is supposed to find the vocabulary id of the masked words. The vocabulary used in BERT is fixed-sized but it consists of subword units called WordPiece instead of full words in a dictionary. That enables the dismantling of input tokens by WordPiece tokenization and thus the handling of unknown words.

After the success of BERT, many variants emerged such as DistilBERT \cite{sanh2019distilbert}, creating a smaller version of BERT with knowledge distillation; ALBERT \cite{lan2020albert}, shrinking BERT by parameter reduction techniques; RoBERTa \cite{liu2019roberta}, optimizing pretraining scheme of BERT for better downstream task performance; XLNet \cite{xlnet-NIPS2019_8812}, highlighting usage of dependency between masked positions; ERNIE \cite{zhang2019ernie}, incorporating knowledge graphs to BERT;  SciBERT \cite{Beltagy2019SciBERT}, pretraining BERT on scientific text; BioBERT \cite{biobert-10.1093/bioinformatics/btz682}, pretraining BERT for biomedicine domain.

DistilBERT \cite{sanh2019distilbert}, which is one of the contextual representations focused in this study along with ELMo \cite{peters-etal-2018-deep}, is built by leveraging knowledge distillation during pretraining of BERT that is 40\% smaller and 60\% faster for restricted training budgets and yet attains 97\% of BERT’s language understanding capabilities. 

\section{Evaluating generalizability}
The evaluation studies before this work are generally designed around a diverse set of downstream tasks to understand knowledge transfer capacity on multiple levels of language input being token, sentence, and document-level \cite{devlin-etal-2019-bert, liu-etal-2019-linguistic, tenney-47786}. The performance is assessed when taken as fixed feature vectors or when the pretrained parameters are further tuned for the tasks at hand \cite{peters-etal-2019-tune}. Ablation studies are also conducted to uncover layer-wise capabilities and effectiveness of different components of the models \cite{liu-etal-2019-linguistic}. Some work focused on how to tune these pretrained representations effectively on sentence and phrase-level text classification \cite{sun-finetune-bert-10.1007/978-3-030-32381-3_16}, named entity recognition, natural language inference, paraphrase detection \cite{peters-etal-2019-tune} to serve as a guide for the NLP practitioner. Howard et al. \cite{howard-ruder-2018-universal} proposed a set of parameter tuning techniques specifically to leverage text classification performance of pretrained language models. Han et al. \cite{han2019unsupervised} applied unsupervised domain adaptation by further pretraining contextual representations on MLM on the target domain. Tenney et al. \cite{tenney-47786} observed that contextual embeddings substantially improve over traditional baselines on learning the syntactic structure of text but that there is only a small improvement in learning semantics on token and sentence level tasks.


\section{Evaluation on multiple contexts}
This thesis study is yet another evaluation study that makes a comparison between two of these representations, namely, ELMo and DistilBERT. But this time, the evaluation is done on part of a recently proposed task set that is realized around a recently proposed news data set: classifying protest news on local news data sets consisting of multiple sentences and coming from different country sources \cite{hurriyetoglu-CLEF}, a cross-country evaluation setting is realized by testing a model on a news text coming from a different country than the training data \cite{hurriyetoglu-CLEF}. Similarly, a sentiment classification task is performed on product reviews, using training and test sets from different domains.

\section{Computational text analysis on contentious politics}
Automating the analysis of text in social science studies is of increasing interest for various reasons. Integrating ML methods with hand labeling can serve as a validation mechanism for human annotations while it also can accelerate the exhaustive analysis of a huge amount of data.

Especially in the case of large-scale studies with a very generic sociological hypothesis to assess, it becomes crucial to gather lots of local information from a diverse set of sources spanning different countries and time \cite{hammond-machine-coded-event, wang-monitoring-societal-events}. Manually labeling each new data becomes infeasible in terms of time and effort. Automating the labeling process, apart from saving many human hours of labeling, can also help standardize annotation procedure by enforcing a control mechanism on human annotations via the model predictions learned by using part of the gold standard data as training set.

Prior work that attempts to automate event coding largely relies on domain-specific rules and keywords that become ineffective outside the target domain. Even in a single domain continuous dictionary update is required that itself can become a burden and also create inconsistencies.

ML methods reduce manual labor dedicated to event coding and eliminate dependence on rules. Neural network-based ML methods, in addition, eliminates most of the feature engineering process. These advantages made researchers turn to automation based on supervision by ML. But one drawback of supervised learning is that it requires lots of labeled data. This poses an ironic contradiction with the ultimate goal of automating information extraction from text, which is the very reason why researchers appeal to ML in the first place. 

The way out of this vicious circle can be found in building generalizable ML systems that can be continuously reused regardless of limitations brought by domain. Also, that kind of a system can play the role of a feedback loop on training-evaluation cycle that provides sufficient dynamism to incorporate new labeled data from different sources much faster.

\section{Cross-context protest news text analysis}
A task set \cite{hurriyetoglu-CLEF} was proposed to collect protest event information from news texts recently to create systems that learn transferable information to extract relevant information from multiple countries with the ultimate motivation to create tools to enable comparative sociology and political studies on social protest phenomena. The task set consists of three tasks: news articles classification, event sentence detection, and event information extraction.

A cross-context setting is realized by providing annotated news from India as train and test sets, and an additional test set that consists of news from China.

In this study, only the first task of the shared task is taken as the probing task for comparison. In this respect, the robustness of ELMo and DistilBERT are evaluated on the binary classification of cross-context protest news articles.

The three tasks mentioned above are realized in the CLEF-2019 Lab ProtestNews on Extracting Protests from News \cite{clef19protest, clef19} in the context of generalizable natural language processing \cite{hurriyetoglu-CLEF}. From the results gathered from 12 teams, it was observed that Neural Networks obtained the best results and a significant drop in cross-country performance is observed on the news from China \cite{Hurriyetoglu-overview-CLEF}. The best performing model on average of null and cross-context trained a BiLSTM with \texttt{fastText} \cite{joulin-etal-2017-fasttext, mikolov2018advances-new-fasttext}
embeddings on a multitask learning objective \cite{radford-clef-levelup}. Safaya et al. \cite{asafaya-clef} attained the smallest score drop between null and cross-contexts using BiGRU and \texttt{word2vec}. Another study utilized ELMo with a fully connected multi layer Neural Network, reaching comparable results \cite{maslennikova-clef}. 

\section{Sentiment Analysis}

Sentiment analysis is a frequently studied classification task. Movie reviews (MR) and customer reviews (CR) are a couple of exhaustively used data sets for this task. As shown in Table \ref{table:sent-sota}, Zhao et al. \cite{Zhao:2015:SHS:2832747.2832816} attained the highest accuracy score 83.1 on MR via feeding 50-dimensional \texttt{word2vec} embeddings to Adasent, a self-adaptive hierarchical sentence representation. InferSent \cite{conneau-etal-2017-supervised}, a sentence representation that is learned by supervised training on a Natural Language Inference data named SNLI \cite{bowman-etal-2015-large}, achieved the best result of 86.3 accuracy on CR \cite{conneau-kiela-2018-senteval}. Logeswaran et al. \cite{logeswaran2018an} achieved accuracy scores of 82.4 for MR and 86.0 for CR by training a multi-channel system consisting of two bi-directional RNNs fed by one fixed and one tunable pretrained word vector representation. These results are followed by skip-thought vectors \cite{kiros-skipthought-NIPS2015_5950}, FastSent, a distributed sentence representation learned from unlabeled data \cite{hill-etal-2016-learning-distributed},  \texttt{fastText} \cite{mikolov2018advances-new-fasttext} and GloVe \cite{pennington-etal-2014-glove} embeddings \cite{conneau-kiela-2018-senteval}. Note that these results are obtained by direct supervision on the tasks by 10-fold cross validation. But  due to time limitations, this study uses a single fold of custom training, development, and test splits. As another difference, here CR results are obtained without any usage of the CR corpus during the training phase.

\begin{table}[H]
\centering
\caption[Sentiment - previous work.]{Sentiment analysis previous work results.}
\vspace{1em}
\begin{tabular}{|l|l|l|}
\hline
\textbf{model}    & \textbf{MR}   & \textbf{CR}   \\ \hline
Zhao et al.\cite{Zhao:2015:SHS:2832747.2832816}       & \textbf{83.1} & 86.1          \\ \hline
Conneau et al. \cite{conneau-etal-2017-supervised}    & 81.1          & \textbf{86.3} \\ \hline
Logeswaran et al. \cite{logeswaran2018an} & 82.4          & 86            \\ \hline
\end{tabular}
\label{table:sent-sota}
\end{table}

\chapter{TASKS}
\label{chapter:tasks}
The transfer capacity of ELMo and DistilBERT are explored under the light of two distinct text classification tasks, each realized under a cross-context experimental setting. One is a document-level binary text classification task that is to classify English news articles from local newspapers of India and China \cite{hurriyetoglu-CLEF}. The other is to classify sentence-level Rotten Tomatoes movie \cite{pang-lee-2005-seeing} and customer reviews \cite{Hu:2004:MSC:1014052.1014073}.

\section{Protest news classification}
For the probing task, two test sets are used: one coming from same corpus as train data (null context), the other coming from a different corpus (cross-context).

The task was designed as an auxiliary task for a currently active research project \cite{hurriyetoglu-CLEF}, whose main motivation is to automate creating news corpus from diverse sources to enable a comparative political and sociological study. Integrating computational linguistics techniques into the pipeline of social studies can considerably accelerate the collection of relevant data. It also enables verification for manual data labeling as a continuous feedback loop. Humanity studies can benefit as much from automation in data collection and analysis \cite{hurriyetoglu-CLEF, sonmez2016towards}. Although, an efficient collection of vast data must be combined with effective labeling. Automation tools must be capable of a certain level of generalization ability to data coming from many different countries, cultures, and eras. Instead of relying on completely domain-specific rules and features, it must be able to also learn generic linguistic features and transfer to new data fast. Otherwise, scalability becomes an issue when incorporating each new data set into the research pipeline.

A shared task set, namely, CLEF-2019 Lab ProtestNews on Extracting Protests from News, was accordingly organized to address the challenge of building NLP tools that are generalizable to different test data \cite{hurriyetoglu-CLEF}. The task set was composed of three levels of news classification tasks: News articles classification, event sentence detection, and event information extraction. A cross-country evaluation setting was realized by training a model on local newspapers of India and testing the model on local newspapers of China. This thesis study borrows the data and the first task setting (news articles classification) as it is.

In this sense, this thesis study differs from the previous work by enforcing a cross-context testing scenario to see how the models generalize to a cross-context test data.

\section{Sentiment analysis}
The second task is the orientation prediction of opinion sentences. Specifically, user reviews on the Internet are marked as ``positive'' or ``negative''. The models are trained and tested on sentence-level MR data \cite{pang-lee-2005-seeing} in the null context, and tested on sentence-level CR data  \cite{Hu:2004:MSC:1014052.1014073} in the cross-context.

Both sentiment data sets were exhaustively used earlier  \cite{kiros-skipthought-NIPS2015_5950, Zhao:2015:SHS:2832747.2832816, conneau-etal-2017-supervised, conneau-kiela-2018-senteval, logeswaran2018an, hill-etal-2016-learning-distributed}. But in none of these studies a cross-context setting is realized. They obtained the result via direct supervision on the target tasks. That is one reason why the results of these work and this thesis study are not comparable. Another factor preventing comparison is that the previous studies apply 10-fold cross-validation to the data.  But  due to time limitations, this study uses a single fold of custom training, development, and test splits.

\chapter{DATA}
\label{chapter:data}

In this study two distinct data sets are used for either task: India and China news articles for PC \cite{hurriyetoglu-CLEF}, movie reviews \cite{pang-lee-2005-seeing} and customer reviews \cite{Hu:2004:MSC:1014052.1014073} for SA. For both tasks, the null context data splits abide by the 75\% - 10\% - 15\% proportions for training, test, and development sets, respectively.

\section{Protest news data}
The news classification task data consists of local India and China news articles in English language. This data set is borrowed from CLEF-2019 Lab ProtestNews \cite{hurriyetoglu-CLEF}. Training, validation, and test splits are provided by the shared task organizers. The news data is specifically a data annotated as to whether it is about a protest event. As illustrated in Table \ref{table:protest-data}, the India data is imbalanced with 22\% protest class, the China data is even more imbalanced with 5\% protest class. 

\begin{itemize}
    \item Local news from India: A news data set in English comprising of news texts gathered from different sources from India which are Times of India, New Indian Express, The Indian Express, and The Hindu.
    \item Local news from China: A news data set in English comprising of news texts of People’s Daily, and South China Morning Post.
\end{itemize}

\begin{figure}[!htbp]
  \centering
  \begin{minipage}[b]{0.45\textwidth}
    \includegraphics[width=\textwidth]{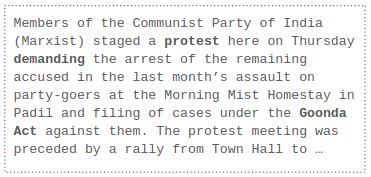}
    \caption[India news sample.]{India news sample.}
    \label{figure:india-sample}
  \end{minipage}
  \hfill
  \begin{minipage}[b]{0.45\textwidth}
    \includegraphics[width=\textwidth]{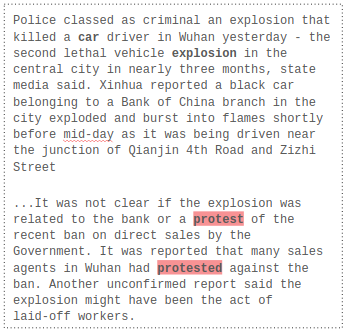}
    \caption[China news sample.]{China news sample.}
    \label{figure:china-sample}
  \end{minipage}
\end{figure}


\subsection{Challenges of political context}
In previous work, it is seen that the classification of contentious political events could be confusing to even domain experts and the inter-annotator agreement could be surprisingly low. That confusion mostly comes from the ambiguity in political terms. How a political event could be interpreted can highly depend on local culture, language usage, time, space and actors. Adding the style and biases of the author of news text, even a single annotator may not be completely sure of herself, let alone agreeing with fellow annotators.

Within the context of contentious politics, ``protest'' can be very broadly defined as engaging in a political dissent via numerous actions such as demonstrating for rights, rallying for political change, conducting a hunger strike, boycotting rights, and so forth.
\subsection{Local news data}
Political events are strongly connected to their local context. Concerning PC, it should be remembered that protest notion might manifest through different kinds of actions in different cultures. In Figure \ref{figure:india-sample}, the news mentions a protest activity as ``Goonda act'', which is a term used in the Indian subcontinent for a hired criminal. In this sense, analyzing local data of many countries can be useful and mostly becomes a necessity to converge to a realistic model of what protest means both globally and locally. 

In this study, ELMo and DistilBERT, which were pretrained on large global and diverse data, encounter with the local news data from two countries. 

\begin{table}[H]
\centering
\caption[Protest - data statistics.]{Protest - data statistics. Ntrain: Training split of India news articles. Ndev: Development split of India news articles. Ntest: Test split of India news articles. Ctest: China news articles as the cross-context test data.}
\vspace{1em}
\begin{tabular}{|l|l|l|l|}
\hline
\textbf{data subset} & \textbf{size} & \textbf{protest ratio}  & \textbf{protest count} \\ \hline
Ntrain              & 3430          & 0.22                    & 754                    \\ \hline
Ndev                & 457           & 0.22                    & 100                    \\ \hline
Ntest               & 687           & 0.22                    & 151                    \\ \hline
Ctest              & 1800          & 0.05                    & 90                      \\ \hline
\end{tabular}
\label{table:protest-data}
\end{table}
\subsection{Long text}
The protest news data set consists of fairly long samples with 300 tokens on average (Here,``token'' is used as a generic term for a unit output of a sequence tokenization process.). This may affect the model performance in two different ways: A model may fail to learn long term relationships within the text. Or a model simply may not be able to utilize the whole text due to memory issues. In this case, very important parts of the data might be lost. For example, the news sample in Figure \ref{figure:china-sample} was indeed falsely labeled as ``non-protest'' by one of the models as the part containing the ``protest'' keyword was clipped.
\subsection{Small data}
Contextual language representations are known to have the potential to substantially reduce the required training data size to create satisfactory models via task-specific fine-tuning on small data. As illustrated in Table \ref{table:protest-data}. the protest news data is also fairly a small one with the number of training samples less than $10000$.

\subsection{Data preparation}
The data preparation process realizes a state-of-the-art semi-automatic pipeline. First, data is randomly sampled from various news sources, and then it is manually annotated by graduate students in the fields related to politics and sociology. A new annotation manual is created based on prominent event annotation guidelines such as ACE \cite{ace2005events} and CAMEO codebook \cite{cameo}. The annotation rules are applied to each data set in the corpus to prevent annotation style ambiguities in error analysis. A supervisor maintains the annotation guideline. The data is annotated in a semi-supervised manner as following: First, students manually annotate by working in pairs, each member of the pairs annotate the same document set on their own. Then the supervisor manually spot-checks 10\% of the agreed annotations and resolves the inter-annotator disagreements. Then a semi-supervised quality check is performed by training an ML model using 90\% of the data as training and validation sets. 10\% of the data as the test set. Training and testing steps are repeated 10 times. Then the errors of the ML model are manually checked.

Using active learning, candidate positive samples are created as follows: Three separate classifiers are trained. If one of the classifiers predicts a document sample as positive, it is sent back to the annotation pipeline. This filtering process runs with  70\% precision and 97\% recall. This procedure aims at building a semi-supervised validation loop to obtain as many relevant documents as possible, without compromising much from precision. It is seen that this technique is much more beneficial in reaching a higher recall of relevant documents than a random sampling of a huge corpus with numerous news topics. Note that for this specific task of protest news collection, recall is a bit more important than precision, for detecting as many relevant articles as possible among numerous other categories is crucial to extract sufficient information to test the hypothesis of the socio-political study. Moreover, recall is critical when creating a gold standard data set.

The China data is used as a cross-context test set for models trained at India data. Any information coming from China data (whether it be the text itself, or the class proportions) is in no way used during training to abide by Protest news Shared Task’s rules for comparability of results with the shared task outcomes.

\section{Sentiment analysis data}
In the second task, movie reviews \cite{pang-lee-2005-seeing} and customer reviews \cite{Hu:2004:MSC:1014052.1014073} data are used. Both data are annotated binary as positive or negative comments. Both data are sentence-level and rather short compared to the protest news data with an average of 20 tokens as illustrated in Table \ref{table:sent-data}. The MR data is class-balanced. There are no predefined training, development, and test splits of both MR and the CR data. Other work using this data generally applies 10-fold cross-validation to these data sets. But this prolongs the hyper-parameter tuning phase to a great extent. Due to time restrictions, a custom split of MR data is created with 75\% training - 10\% development - 15\% test proportions. CR data is used as the cross-context test set as is.

\begin{table}[H]
\centering
\caption[Sentiment data samples.]{Sentiment data samples.}
\vspace{1em}
\begin{tabular}{|l|l|l|}
\hline
\textbf{data} & \textbf{examples}                                              & \textbf{label} \\ \hline
MR            & “Too slow for a younger crowd , too shallow for an older one.” & neg            \\ \hline
CR            & “We tried it out christmas night and it worked great .”        & pos            \\ \hline
\end{tabular}
\label{table:sent-data}
\end{table}

\begin{table}[H]
\centering
\caption[Sentiment - data statistics.]{Sentiment - data statistics. Ntrain: Training split of MR data set. Ndev: Development split of MR data set. Ntest: Test split of MR data set. Ctest: CR data set as the cross-context test data.}
\vspace{1em}
\begin{tabular}{|l|l|l|l|l|}
\hline
\textbf{data subset} & \textbf{size} & \textbf{positive ratio}  & \textbf{avg len} & \textbf{stdev} \\ \hline
Ntrain               & 7974          & 0.5                                       & 21               & 9.44           \\ \hline
Ndev                 & 1088          & 0.5                                        & 21               & 9.3            \\ \hline
Ntest                & 1600          & 0.5                                        & 20               & 9.34           \\ \hline
Ctest                & 3771          & 0.64                                      & 20               & 11.71          \\ \hline
\end{tabular}
\label{table:sent-data}
\end{table}
\chapter{EXPERIMENTAL SETUP}
\label{chapter:experimental-setup}
\section{Environment setup}
The classifiers are implemented in Python programming language using the PyTorch library \cite{pytorch}. The training is done on a single V100 NVIDIA GPU with 16 GB RAM.

\section{ELMo}

ELMo \cite{peters-etal-2018-deep} is a deep context-dependent representation learned from the internal states of a deep BiLM that is acquired by the joint training of two LSTM layers on both directions. ELMo obtained state-of-the-art task-specific supervised models in six different NLP tasks, including sequence classification. It is shown that ELMo's every BiLM layer represents a different type of information, together capturing syntax and disambiguating semantics.

ELMo is pretrained for BiLM objective on One Billion Words Benchmark \cite{onebillion} composed of nearly 30 million monolingual text data obtained from the WMT11 website \cite{wmt11-text}. ELMo word embeddings are created using two BiLM layers and make us of character convolutions, which makes it robust to out-of-vocabulary tokens unseen in training.

This study makes use of the original pretrained ELMo model with 2 layer bidirectional LSTM layers with 4096 units and 512-dimensional projections, with a total of 93.6 million parameters.
ELMo’s hidden LSTM layers are weighted averaged (scalar-mix method), which is the default combination technique of ELMo, and then fed into the classifier layers.
\section{DistilBERT}
DistilBERT \cite{sanh2019distilbert} is created by applying knowledge distillation to BERT, specifically the \texttt{bert-base-uncased} model, to create a lighter and faster version of BERT for the benefit of under constraint training budgets. It is 40\% smaller and 60\% faster than the respective BERT model while attaining 97\% of BERT's performance on 6 of 12 downstream tasks. Only the MLM is used as the pretraining objective, omitting BERT's other objective Next Sentence Prediction. The model was trained on the concatenation of English Wikipedia and Toronto BookCorpus \cite{bookcorpus}. To create a smaller version of BERT, DistilBERT's creators removed the token-type embeddings and the pooler from the architecture and reduced the number of layers by a factor of 2. In this study, DistilBERT’s last four hidden layers are simply averaged and fed into the classifier layers, which is a suggested usage of BERT for text classification tasks. 

The reason why DistilBERT is preferred over BERT is that training BERT gives out-of-memory errors in a single GPU with 256 tokens per sample (even when frozen), whereas DistilBERT does not.

Knowledge distillation is a neural network compression technique in which a pretrained larger model is mimicked by a smaller model \cite{bucila-distil}. The larger model is referred as ``teacher'', the smaller model is referred as ``student''. The student model is trained against the weighted sum of two kinds of losses: ``distillation loss'' and ``student loss''. Student loss is just the usual cross-entropy loss calculated by putting class probabilities predicted by the student model against the hard labels of the training data. ``Distillation loss'' is found by comparing the class probability predictions of the student and teacher (Note that teacher is a pretrained model and its predictions are ready at hand). So the weighted sum of these two losses are tried to be minimized during training.

A Neural Network's original class probabilities typically follow this pattern regarding the class probability values: A probability that is close to one is assigned to the predicted class. The remaining class probabilities are close to zero. This situation makes teacher's class probabilities very similar to the hard labels. Hinton et al. \cite{hinton-distil} emphasize that the models actually do not favor one class over the other classes that strongly. In reality, some class probabilities are closer to the favored class probability. To uncover this ``dark knowledge'', Hinton et al. \cite{hinton-distil} smoothed the class probabilities by dividing the logits by an integer value called `softmax temperature''. That means, the teacher probabilities used in ``distillation loss'' are smoothed by reorginizing the softmax formula as follows:
\begin{equation}
    p_{i} = \frac{exp(\frac{z_{i}}{T})}{\sum_{j}exp(\frac{z_{j}}{T})}
\end{equation}
where $p_{i}$ is the probability for class $i$, $z_{i}$ is the logit value for class $i$, and $T$ is the softmax temperature (generally varying between 1 and 20). Note that when $T=1$, the formula equals to the regular softmax function.

So the distillation loss can be formulated as follows \cite{kd-nervana}:
\begin{equation}
    L(x;W) = \alpha*H(y,\sigma(z_{s};T=1)) + \beta*H(\sigma(z_{t};T=\tau), \sigma(z_{s};T=\tau))
\end{equation}
where $x$ is the input, $y$ is the ground truth label, $W$ are the student model parameters, $H$ is the loss function, $\sigma$ is the softmax function, $T$ is the softmax temperature, $z_{s}$ and $z_{t}$ are logits of the student and teacher models, respectively. The two losses are weighted via $\alpha$ and $\beta$ parameters.

In this study, \texttt{distilbert-base-uncased} \cite{distilbert} with 66 million parameters is compared to the original ELMo model with 93.6 million parameters \cite{elmo}. Note that while the former is case-insensitive, the latter is case-sensitive. Also, they were pretrained on different unlabeled data. In this sense, the comparison must be viewed as realized not just between the models, but rather between model-unlabeled pretraining data pairs.

\section{Baseline models}
Optimized Linear SVM (LSVM) and Multinomial Naive Bayes (MNB) scores are reported as baselines \cite{scikitlearn}. LSVM takes the input as tf-idf (term frequency - inverse document frequency) vectors, whereas MNB as a sparse vector of token counts. 

Sparse vector of token counts is a vector of type Bag of Words which stores number of occurrences in a document for each word in the vocabulary ($V$). It is a ``sparse'' representation since each vector is of size $|V|$ and the vectors are dominated by the number ``zero'' because each document typically contains only a small subset of the words in the vocabulary. 

Tf-idf vectors are also of size $|V|$, but this time, instead of counts, the vectors keep a continuous value for each word to refer to a word's ``importance'' in that document. According to this representation, the ``importance'' of a word is proportional to its frequency in a document, and disproportional to its frequency across all documents in the data set. In this sense, tf-idf score for a word is calculated by 
\begin{equation}
    w_{i,j} = tf_{i,j} * log(\frac{N}{df_{i}})
\end{equation}
where $i$ is the word's index in the vocabulary, $j$ is the document's index, $N$ is the number of documents in the data set, $df_{i}$ is the document frequency of the word $i$. The expression $log(\frac{N}{df_{i}})$ equals to the inverse document frequency. Term frequency in turn is calculated as 
\begin{equation}
    tf_{i,j}=\frac{n_{i,j}}{\sum_{k}n_{i,j}}
\end{equation}
where the count of the word $i$ in document $j$ is normalized by document length (number of words in document $j$). 

It is important to see how efficient traditional ML algorithms in size and speed compared to heavily pretrained large contextual networks. This enables to understand if the overhead of the deep contextual models is worth to undertake. Note that the baseline models are much more simpler than the neural classifiers described in the Figure \ref{figure:classarch}. Moreover, here the baseline models utilize simple word representations that are tf-idf and count vectors which do not preserve word order and context information.

\section{Classifiers and training schedule}
\begin{figure}[!htbp]
  \centering
  \includegraphics[width=0.45\textwidth, height=0.5\textheight]{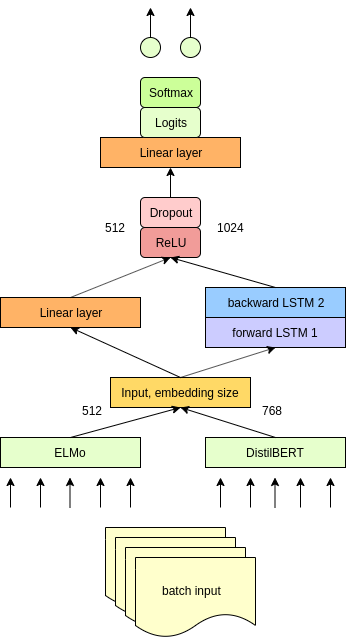}
    \caption[Classifier architecture.]{Classifier architecture. BiLSTM classifier (right side) is created by simply replacing the bottom linear layer of the FFNN (left side) with a 2-layer BiLSTM.}
    \label{figure:classarch}
\end{figure}

The classifier architectures are kept simple to focus on what information can be easily extracted from ELMo and DistilBERT. First, a 2-layer FFNN with 512 hidden units is used. Then, to better understand the effect of adding task-trained contextualization, a 2-layer BiLSTM with 512 hidden units is added before the linear output layer (See Figure \ref{figure:classarch}). The default maximum sequence length is 256 tokens for PC, 60 tokens for SA. ELMo gets that many full tokens, whereas DistilBERT gets that many WordPiece outputs.

The classifiers are trained for 10 epochs with Adam optimizer \cite{adam-Kingma} using step decay with the patience of 3 epochs on the development set F-score. If F-score on development set does not improve throughout 3 consecutive epochs, the learning rate is dropped by the decay rate that was determined in hyper-parameter-tuning phase. Step decay is used to prevent the model from overfitting the training data. 

One common method to detect overfitting in neural networks is to track the score on the development set. A halt or a decrease in development score improvement is an indicator to overfitting. Early stopping is an alternative technique to learning rate to prevent overfitting, which is simply the termination of training. In this study step decay is preferred over early stopping to allow models to benefit equal number of epochs. Step decay is preferred as the learning rate annealing method based on an online course's recommendation \cite{cs231n}. 

During training, the best model is checkpointed regarding the development set F-score. Then the checkpoints are evaluated on the test data. This procedure is repeated for each classifier with 5 random seeds and the average scores are reported. 

Except for DistilBERT, the sequences are tokenized by Spacy's \texttt{en-core-web-sm} tokenizer\cite{spacy}. DistilBERT uses WordPiece tokenization \cite{45610}. The first 256 tokens per sample of the protest news data and first 60 tokens per sample of the MR and CR data are given as input to the classifiers. Note that the usage of two different tokenizers causes a mismatch between the input of DistilBERT and other models. But WordPiece tokenization is preferred for DistilBERT as it is the default tokenizer of it.

\section{Experiments}
\begin{figure}[!htbp]
  \centering
   \includegraphics[width=0.8\textwidth]{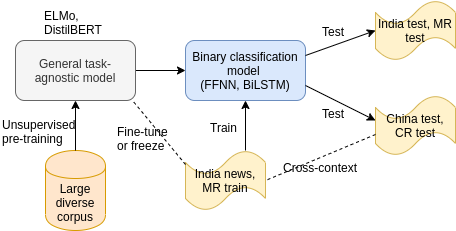}
    \caption[Experimental setup.]{Experimental setup.}
    \label{figure:expsetup}
\end{figure}

The main focus of this study is to acquire an idea of the robustness quality of ELMo and DistilBERT under the usage of cross-context test data (See Figure \ref{figure:expsetup}). Observing both null and cross-context performance side-by-side helps understand cross-context robustness more easily. Therefore, all experiments report both null and cross-context results for each task. Each experiment focuses on a particular variation on the classifier architecture that possibly affects the results in its way.
\begin{itemize}
    \item Experiment 1: Frozen embeddings: Both contextual word embeddings are used as fixed word vectors and fed into FFNN.
    \item Experiment 2: Fine-tuned embedding: ELMo and DistilBERT are fine-tuned to the training data sets together with the FFNN classifier. 
    \item Experiment 3: Using BiLSTM: Both models are kept frozen, but this time paired with a BiLSTM instead of an FFNN.
    \item Experiment 4: Fine-tuning + BiLSTM: ELMo and DistilBERT are compared under the combined effect of fine-tuning contextual embeddings and pairing with a 2-layer BiLSTM.
\end{itemize}

\section{Evaluation criteria}
The protest news data is highly imbalanced regarding the class frequencies (negative class is dominant). Macro-averaging provides a more robust evaluation for class-imbalanced data. On the other hand, macro-averaging and micro-averaging are equivalent in the case of class-balanced data sets, such as the MR data. Macro averaged F-score ($\beta=1$) is used as the primary evaluation metric in both tasks. To capture cross-context robustness better, also as an additional metric, the F-score drop between null and cross-context is tracked in percentages. That is, for example, if model $x$ null context F-score is $f_{n}$ and its cross-context setting F-score is $f_{c}$, then the drop in F-score is calculated as $(f_{n}-c_{n})/f_{n}*100$. Since PC data is imbalanced, positive class scores are also recorded in PC.  Accuracy in SA is additionally reported for compatibility with the previous work in Appendix \ref{chapter:det-res} as well as standard deviations.


So three evaluation scores are recorded for each model for both tasks:
\begin{itemize}
    \item F-score in the null context (Ntest).
    \item F-score in the cross-context (Ctest).
    \item The F-score drop between null and cross-context tracked in percentages (Drop).
\end{itemize}

The other metrics reported for PC;
\begin{itemize}
    \item  F-score for positive  (protest) class in null context (Npos).
    \item  F-score for positive class in cross-context (Cpos).
    \item  The F-score drop for positive class between null and cross-context tracked in percentages (Drop pos).
\end{itemize}

Positive class F-score is calculated by taking the harmonic mean of positive class precision and positive class recall. Positive class precision is found as $\frac{tp}{pp}$ and positive class recall is found as $\frac{tp}{ap}$ where $tp$ is the number of instances that are truly predicted as positive class, $pp$ is the number of instances ``predicted as positive'' by the classifier, and $ap$ is the number of instances that are ``actually positive''. 

In addition, the average of null and cross context F-score is also reported (``Test avg''). This metric enables to distinguish between performances of the models on the average. Taking the average score into account can be useful especially when a model's ``Drop'' is low (which is desired) but the same model's average performance is actually rather low. In this case, another model that performs better on average could be favored, even if its ``Drop'' is higher than the other. The average for the positive class scores is also reported (``Pos avg'').

\section{Hyper-parameter tuning} 
The dropout rate of the classifier (FFNN or BiLSTM), learning rate, learning rate decay, L2 norm, and whether to use ReLU or not, are the hyper-parameters that were tuned for each model. Hyper-parameter space to traverse is illustrated in Table \ref{table:hyp-space} that is determined within the range of the hyper-parameter values settled in related work \cite{peters-etal-2018-deep, devlin-etal-2019-bert, sanh2019distilbert, liu-etal-2019-linguistic, peters-etal-2019-tune, sun-finetune-bert-10.1007/978-3-030-32381-3_16}. Hyper-parameters of ELMo and DistilBERT are kept unchanged.

\begin{table}[H]
\centering
\caption[Hyper-parameter space.]{Hyper-parameter space.}
\vspace{1em}
\begin{tabular}{|l|l|}
\hline
\textbf{hyper-parameter} & \textbf{range}   \\ \hline
learning rate           & 5e-5, 1e-3, 1e-1 \\ \hline
learning rate decay     & 0, 0.5           \\ \hline
dropout                 & 0, 0.25, 0.5     \\ \hline
L2                      & 0, 0.01          \\ \hline
Use ReLu?               & True, False      \\ \hline
\end{tabular}
\label{table:hyp-space}
\end{table}

Hyper-parameters of each distinct model is optimized on the validation data with The Tree-structured Parzen Estimator (TPE) algorithm \cite{Bergstra:2011:AHO:2986459.2986743}. The implementation of the algorithm is provided by the hyperopt package \cite{Bergstra:2013:MSM:3042817.3042832}. TPE mainly traverses a hyper-parameter space by pruning it effectively to optimize a predefined metric on the validation subset of the data. TPE is shown to converge faster to the best hyper-parameters than random or grid search \cite{Bergstra:2011:AHO:2986459.2986743}.

In SA task, cross-validated data required nested cross-validation during hyper-parameter tuning, which can get out of hand if hyper-parameter space contains more than a couple of points. But in this study, the hyper-parameter space contains 72 different combinations of points for the neural networks. This number is reduced to 25 with a bayesian-based tuning technique. This process has to be done for 8 different models (4 ELMo, 4 DistilBERT) for SA. There was not much time left for that kind of tuning. Therefore, a custom stratified training, development, and test splits of the MR data is created and the hyper-parameter tuning was performed on that single fold which is used throughout the experiments.

\begin{table}[H]
\centering
\caption[Hyper-parameter space - baselines.]{Hyper-parameter space of the baselines.}
\vspace{1em}
\begin{tabular}{|l|l|l|}
\hline
\textbf{model} & \textbf{hyper-parameter} & \textbf{range}        \\ \hline
MNB            & alpha                    & 0, 0.25, 0.5, 0.75, 1 \\ \hline
MNB            & fit-prior                & True, False           \\ \hline
LSVM           & loss                     & hinge, squared hinge  \\ \hline
LSVM           & tolerance                & 1e-2, 1e-3, 1e-4      \\ \hline
LSVM           & C                        & 0.5, 1                \\ \hline
\end{tabular}
\label{table:hyp-base}
\end{table}

The hyper-parameter tuning for the baselines was straightforward, for there were a few possible hyper-parameters to be tuned as seen in Table \ref{table:hyp-base}. For MNB, ``alpha'' the additive smoothing parameter, and ``fit-prior'', the choice of whether to learn class prior probabilities or not, are tuned. For LSVM, the loss function type as ``hinge'' or ``squared hinge'', the tolerance for stopping criteria, and ``C'' the regularization parameter are tuned.

\section{Statistical tests}
Two statistical tests are performed on ELMo and DistilBERT results: randomization test and McNemar's test. The former checks if the two models significantly differ in terms of scores. The latter checks if they differ in their failures in predicting the same samples.

\subsection{Randomization test}
The first statistical test is the randomization (permutation) test \cite{Yeh-randtest}, which is a technique to see if the difference between the performance of two models on some metric is significant or happened by chance. Randomization test is favorable in NLP problems over other widely used statistical test techniques such as the t-test. Because, unlike the t-test, the randomization test does not assume any distribution over data. And it also does not assume independence between predictions of two models. 

Randomization test is performed by calculating p-values for all combinations of predictions obtained by training with different seeds. For example, when two models of ELMo and DistilBERT are compared, 25 different p-values are produced by using 25 different pairs of 5 ELMo and 5 DistilBERT outcomes. The harmonic mean of these p-values is used as the ultimate statistic of the test to smooth the disproportional effect of large p-values occuring in arithmetic mean. 


The harmonic mean of a series equals to zero if the series contains any zero value. For more realistic evaluation, the harmonic mean of non-zero p-values is also reported. But zero values should not be entirely ignored since their existence points out that rejection of the null hypothesis is indeed very much probable.

As a widely used technique, the matched-pair t-test is not an equally suitable option for two reasons: Firstly, the matched-pair t-test assumes a Gaussian distribution over the statistical value it uses. Moreover, the matched-pair t-test is effective when comparing recall, but not as effective when comparing precision and F-score. This is because precision and F-score are more complicated and non-linear functions of random variables than recall \cite{Yeh-randtest}. 

\subsection{McNemar's test}
The other statistical test applied here is McNemar’s test which is proved to have low Type I error \cite{Dietterich:1998:AST:303222.303237}. Unlike the randomization test, McNemar's test is interested in how the modes label specific test instances. McNemar’s test aims at measuring the level of discrepancy between the models' failures.
The overall p-value for a single test is calculated in the same way as the randomization test (taking harmonic mean).

\chapter{EXPERIMENT RESULTS}
\label{chapter:experiment-results}
In this section, ELMo and DistilBERT are compared using various classification architectures on two text classification tasks in a cross-context.
\section{Comparing ELMo and DistilBERT}
This section consists of direct comparison of ELMo and DistilBERT by isolating them under different circumstances enabled by varying the training style (freeze or fine-tune embeddings),  and the classifier architecture (adding BiLSTM to the classifier). 

As a side note, fine-tuned ELMo does not fit into a single GPU with 256 tokens per input sample. This issue is encountered in PC, for news inputs consist of multiple sentences. In this case, when fine-tuned, ELMo could manage up to 150 tokens per input. For a fair comparison, DistilBERT is also fine-tuned with 150 tokens. But since DistilBERT can handle 256 tokens per input, DistilBERT results on 256 tokens per input are also reported. 

\subsection{Experiment 1: Frozen embeddings}
In the first experiment, both contextual word embeddings are used as fixed word vectors without further fine-tuning on the two classification tasks and fed into FFNN. In this sense, here what is compared is the generalization capacity of the models coming only from the pretraining phase.

The numbers ``256'' and ``60'' next to model names indicate the number of tokens used per training, validation, and test instances.

Table \ref{table:exp1-pc} and \ref{table:exp1-sa} show that frozen DistilBERT is on par with or exceeding frozen ELMo in the null context (India and MR test sets). On the other hand, as illustrated in the ``Drop'' column in Table \ref{table:exp1-pc}, DistilBERT outperforms ELMo in the cross-contexts with a smaller ``Drop'' score in both tasks.

\begin{table}[H]
\centering
\caption[Protest - frozen embeddings.]{PC results of frozen ELMo and DistilBERT combined with FFNN. }
\vspace{1em}
\begin{adjustbox}{width=1\textwidth}
\begin{tabular}{|l|l|l|l|l|l|l|l|l|l|}
\hline
\textbf{task} & \textbf{model} & \textbf{Ntest} & \textbf{Npos} & \textbf{Ctest} & \textbf{Cpos} & \textbf{Test avg} & \textbf{Pos avg} & \textbf{Drop} & \textbf{Pos Drop} \\ \hline
PC       & ELMo 256       & 83.6           & 74.8          & 75.2           & 52.6          & 79.4              & 63.7             & 10            & 29.6              \\ \hline
PC       & DistilBERT 256 & \textbf{83.8}  & \textbf{75}   & \textbf{76.8}  & \textbf{56}   & \textbf{80.3}     & \textbf{65.5 }            & \textbf{8.2}          & \textbf{25.2}              \\ \hline
\end{tabular}
\end{adjustbox}
\label{table:exp1-pc}
\end{table}

\begin{table}[H]
\centering
\caption[Sentiment - frozen embeddings.]{SA results of frozen ELMo and DistilBERT combined with FFNN.}
\vspace{1em}
\begin{tabular}{|l|l|l|l|l|l|}
\hline
\textbf{task} & \textbf{model} & \textbf{Ntest} & \textbf{Ctest} & \textbf{Test avg} & \textbf{Drop} \\ \hline
SA     & ELMo 60        & 78             & 63.6           & 70.8              & 18.4          \\ \hline
SA     & DistilBERT 60  & \textbf{79}    & \textbf{66.8}  & \textbf{72.9}     & \textbf{15.4} \\ \hline
\end{tabular}
\label{table:exp1-sa}
\end{table}

\subsection{Experiment 2: Fine-tuned embeddings}
In the second part of the experiments, ELMo and DistilBERT are fine-tuned on the training data sets together with the FFNN classifier. 

For PC, fine-tuned ELMo, does not fit into a single GPU with 256 tokens per input sample. This issue is encountered in PC, for news inputs consist of multiple sentences. In this case, ELMo could manage up to 150 tokens per input. For a fair comparison, DistilBERT is trained twice, first with 150 tokens of input, and then as a separate model, with 256 tokens of input. The maximum sequence limit does not become an issue for SA, as the sentiment data is sentence-level and much shorter (300 tokens vs. 20 tokens on average).

\begin{table}[H]
\centering
\caption[Protest - fine-tuned embeddings.]{PC results of fine-tuned ELMo and DistilBERT combined with FFNN.}
\vspace{1em}
\begin{adjustbox}{width=1\textwidth}
\begin{tabular}{|l|l|l|l|l|l|l|l|l|l|}
\hline
\textbf{task} & \textbf{model}    & \textbf{Ntest} & \textbf{Npos} & \textbf{Ctest} & \textbf{Cpos} & \textbf{Test avg} & \textbf{Pos avg} & \textbf{Drop} & \textbf{Pos Drop} \\ \hline
PC       & ELMo ft 150       & 83             & 74            & 72.2           & 47            & 77.6              & 60.5             & 12.8          & 36.4              \\ \hline
PC       & DistilBERT ft 150 & 80             & 69            & 71             & 45.6          & 75.5              & 57.3             & 11            & 33.8              \\ \hline
PC       & DistilBERT ft 256 & \textbf{83.2}  & \textbf{74.6} & \textbf{76.4}  & \textbf{55.4} & \textbf{79.8}     & \textbf{65}      & \textbf{8.2}  & \textbf{25.6}     \\ \hline
\end{tabular}
\end{adjustbox}
\label{table:exp2-pc}
\end{table}

\begin{table}[H]
\centering
\caption[Sentiment -  fine-tuned embeddings.]{SA results of fine-tuned ELMo and DistilBERT combined with FFNN.}
\vspace{1em}
\begin{tabular}{|l|l|l|l|l|l|}
\hline
\textbf{task} & \textbf{model}   & \textbf{Ntest} & \textbf{Ctest} & \textbf{Test avg} & \textbf{Drop} \\ \hline
SA     & ELMo ft 60       & 76.2           & \textbf{69}    & 72.6              & \textbf{9.6}  \\ \hline
SA     & DistilBERT ft 60 & \textbf{79}    & 68             & \textbf{73.5}     & 14.2          \\ \hline
\end{tabular}
\label{table:exp2-sa}
\end{table}

As illustrated in Table \ref{table:exp2-pc}, when fine-tuned, ELMo outperforms DistilBERT when the context is restricted to 150 tokens, but falls behind in 256 tokens especially in cross-context. Table \ref{table:exp2-sa} shows that DistilBERT surpasses ELMo in the null context, but falls behind in the cross-context. This indicates that in SA, fine-tuning made ELMo more robust to context change in the test set. 

Here it should be noted that DistilBERT does not use as many full tokens as Elmo since DistilBERT adopts WordPiece tokenization on default. Given this situation, providing DistilBERT as many full tokens would have allowed a fairer comparison. Though in this study, this modification is not done to minimize the effect of any customization upon functions of ELMo and DistilBERT’s libraries, including tokenization style. 

\subsection{Experiment 3: External contextualization via BiLSTM}

In this experiment, both models are kept frozen, but this time paired with a BiLSTM instead of an FFNN. BiLSTM adds contextualization on the focused task, thus it is expected to improve results.

\begin{table}[H]
\centering
\caption[Protest - frozen with BiLSTM.]{PC results of frozen ELMo and DistilBERT combined with BiLSTM.}
\vspace{1em}
\begin{adjustbox}{width=1\textwidth}
\begin{tabular}{|l|l|l|l|l|l|l|l|l|l|}
\hline
\textbf{task} & \textbf{model}          & \textbf{Ntest} & \textbf{Npos} & \textbf{Ctest} & \textbf{Cpos} & \textbf{Test avg} & \textbf{Pos avg} & \textbf{Drop} & \textbf{Pos Drop} \\ \hline
PC       & ELMo + BiLSTM 256       & 81.6           & 72.2          & 72.4           & 47.6          & 77                & 59.9             & 11.2          & 34                \\ \hline
PC       & DistilBERT + BiLSTM 256 & \textbf{84.2}  & \textbf{76.4} & \textbf{78.4}  & \textbf{59}   & \textbf{81.3}     & \textbf{67.7}    & \textbf{7}    & \textbf{23}       \\ \hline
\end{tabular}
\end{adjustbox}
\label{table:exp3-pc}
\end{table}

\begin{table}[H]
\centering
\caption[Sentiment - frozen with BiLSTM.]{SA results of frozen ELMo and DistilBERT combined with BiLSTM.}
\vspace{1em}
\begin{tabular}{|l|l|l|l|l|l|}
\hline
\textbf{task} & \textbf{model}         & \textbf{Ntest} & \textbf{Ctest} & \textbf{Test avg} & \textbf{Drop} \\ \hline
SA     & ELMo + BiLSTM 60       & 79             & 67             & 73                & 15.2          \\ \hline
SA     & DistilBERT + BiLSTM 60 & \textbf{80}    & \textbf{70.2}  & \textbf{75.1}     & \textbf{12.4} \\ \hline
\end{tabular}
\label{table:exp3-sa}
\end{table}

As the Tables \ref{table:exp3-pc} and \ref{table:exp3-sa} illustrate, in both tasks DistilBERT outperforms ELMo when paired with a 2-layer BiLSTM. The gap is more visible in the cross-context performance: As shown in Table \ref{table:exp3-pc}, DistilBERT surpasses ELMo $6$ points with a 78.4 Ctest F-score.

\subsection{Experiment 4: Combining fine-tuning with BiLSTM}
In this experiment, ELMo and DistilBERT are compared under the combined effect of fine-tuning and the usage of 2-layer BiLSTM. Note that, in PC, ELMo could handle at most 150 tokens per input. Therefore, the comparison is done under that much of a sequence length.

\begin{table}[H]
\centering
\caption[Protest - fine-tuned with BiLSTM.]{PC results of fine-tuned ELMo and DistilBERT combined with BiLSTM.}
\vspace{1em}
\begin{adjustbox}{width=1\textwidth}
\begin{tabular}{|l|l|l|l|l|l|l|l|l|l|}
\hline
\textbf{task} & \textbf{model}             & \textbf{Ntest} & \textbf{Npos} & \textbf{Ctest} & \textbf{Cpos} & \textbf{Test avg} & \textbf{Pos avg} & \textbf{Drop} & \textbf{Pos Drop} \\ \hline
PC       & ELMo ft + BiLSTM 150       & \textbf{82}    & \textbf{72.4} & 72             & 46.6          & \textbf{77}       & 59.5             & 12.2          & 35.6              \\ \hline
PC       & DistilBERT ft + BiLSTM 150 & 81.8           & 71.8          & \textbf{72.2}  & \textbf{47.6} & \textbf{77}       & \textbf{59.7}    & \textbf{11.8} & \textbf{33.6}     \\ \hline
\end{tabular}
\end{adjustbox}
\label{table:exp4-pc}
\end{table}

\begin{table}[H]
\centering
\caption[Sentiment - fine-tuned with BiLSTM.]{SA results of fine-tuned ELMo and DistilBERT combined with BiLSTM.}
\vspace{1em}
\begin{tabular}{|l|l|l|l|l|l|}
\hline
\textbf{task} & \textbf{model}            & \textbf{Ntest} & \textbf{Ctest} & \textbf{Test avg} & \textbf{Drop} \\ \hline
SA     & ELMo ft + BiLSTM 60       & 78.2           & 67.4           & 72.8              & 13.8          \\ \hline
SA     & DistilBERT ft + BiLSTM 60 & \textbf{80}    & \textbf{70.2}  & \textbf{75.1}     & \textbf{12.4} \\ \hline
\end{tabular}
\label{table:exp4-sa}
\end{table}

As a reminder, in Experiment 2, DistilBERT was underperforming on sequences of length 150 in PC. Now, as illustrated in Table \ref{table:exp4-pc} DistilBERT catches up with ELMo. This indicates that DistilBERT benefits from BiLSTM. Again, if Experiment 2 is recalled, DistilBERT was occasionally underperforming ELMo in SA. But now the results in Table \ref{table:exp4-sa} are in favor of DistilBERT. This indicates BiLSTM enabled DistilBERT to surpass ELMo results.

\subsection{Comparison to baselines}
For fairness, both ELMo’s and DistilBERT’s best and worst-performing configurations are compared to the hyper-parameter-tuned MNB and LSVM baselines. The best performing models are indicated with the keywords ``highest'', the worst-performing with ``lowest'' on Table \ref{table:comp-baseline-pc} and \ref{table:comp-baseline-sa}. Two models are reported as the ``highest'' of ELMo in SA as one owns better ``Drop'' scores.

\begin{table}[H]
\centering
\caption[Protest - baselines.]{PC - Comparison with the baselines.}
\vspace{1em}
\begin{adjustbox}{width=1\textwidth}
\begin{tabular}{|l|l|l|l|l|l|l|l|l|l|}
\hline
\textbf{task} & \textbf{model}                       & \textbf{Ntest} & \textbf{Npos} & \textbf{Ctest} & \textbf{Cpos} & \textbf{Test avg} & \textbf{Pos avg} & \textbf{Drop} & \textbf{Pos Drop} \\ \hline
PC       & LSVM 256                       & 79             & 67            & 64             & 30            & 71.5              & 48.5               & 19            & 55                \\ \hline
PC       & MNB 256                              & 80             & 71            & 73             & 49            & 76.5              & 60               & 9            & 31                \\ \hline
PC       & ELMo + BiLSTM 256 (lowest)           & 81.6           & 72.2          & 72.4           & 47.6          & 77                & 59.9             & 11.2          & 34                \\ \hline
PC       & ELMo 256 (highest)                   & 83.6           & 74.8          & 75.2           & 52.6          & 79.4              & 63.7             & 10            & 29.6              \\ \hline
PC       & DistilBERT ft 256 (lowest)           & 83.2           & 74.6          & 76.4           & 55.4          & 79.8              & 65               & \textbf{8.2}  & 25.6              \\ \hline
PC       & DistilBERT 256 (highest) & \textbf{83.8}  & \textbf{75}   & \textbf{76.8}  & \textbf{56}   & \textbf{80.3}     & \textbf{65.5}    & \textbf{8.2}  & \textbf{25.2}     \\ \hline
\end{tabular}
\end{adjustbox}
\label{table:comp-baseline-pc}
\end{table}

\begin{table}[H]
\centering
\caption[Sentiment- baselines.]{SA - Comparison with the baselines.}
\vspace{1em}
\begin{tabular}{|l|l|l|l|l|l|}
\hline
\textbf{task} & \textbf{model}                      & \textbf{Ntest} & \textbf{Ctest} & \textbf{Test avg} & \textbf{Drop} \\ \hline
SA     & MNB 60                              & 78             & 57             & 67.5                & 27            \\ \hline
SA     & LSVM 60                       & 77             & 62             & 69.5              & 19            \\ \hline
SA     & ELMo 60 (lowest)                    & 78             & 63.6           & 70.8                & 18.4          \\ \hline
SA     & ELMo ft 60 (highest 1)              & 76.2           & 69             & 72.6              & \textbf{9.6}  \\ \hline
SA     & ELMo + BiLSTM 60 (highest 2)        & 79             & 67             & 73                & 15.2          \\ \hline
SA     & DistilBERT 60 (lowest)              & 79             & 66.8           & 72.9                & 15.4          \\ \hline
SA     & DistilBERT ft + BiLSTM 60 (highest) & \textbf{80}    & \textbf{70.2}  & \textbf{75.1}     & 12.4          \\ \hline
\end{tabular}
\label{table:comp-baseline-sa}
\end{table}

Table \ref{table:comp-baseline-pc} demonstrates that in PC, while LSVM cannot catch up with any model, MNB performs fairly on par with ELMo’s worst-performing model. Apart from that, MNB is effectively surpassed by the best of ELMo and DistilBERT in all tracks. As shown in Table \ref{table:comp-baseline-sa}, in SA, MNB is inferior to all models. The results of LSVM and ELMo's lowest are close to each other. But the best of ELMo and all variants of DistilBERT surpass the LSVM baseline with an apparent gap in the cross-context robustness.

It is also visible in the Tables \ref{table:comp-baseline-pc} and \ref{table:comp-baseline-sa} that DistilBERT outperforms ELMo on both tasks with both of its worst-performing and best-performing variants. This can be viewed as an indicator of the possible superiority of DistilBERT.

\subsection{Average results}
To view the experiment from a wider perspective, the models are also compared under the arithmetic average of all variations. For fairness, two different average values are reported for  DistilBERT: 1) the average of variations that are also ``common'' to ELMo. 2) the average over DistilBERT's all variations. In the first average, DistilBERT’s fine-tuned models making use of 256 length input are excluded from the computation because there is no equivalent model on the ELMo side. 

As Table \ref{table:avg-results-pc} displays, ELMo is found to be superior to DistilBERT on average when both use only 150 tokens of protest news input. But as illustrated in Table \ref{table:avg-results-sa}, in SA when full context is available DistilBERT performs better regardless of short sequence length. On average of common variations, DistilBERT is dominant in both tasks. This can be seen as an indicator of DistilBERT's overall superiority.

\begin{table}[H]
\centering
\caption[Protest - average scores.]{PC - Average results of ELMo and DistilBERT.}
\vspace{1em}
\begin{adjustbox}{width=1\textwidth}
\begin{tabular}{|l|l|l|l|l|l|l|l|l|l|}
\hline
\textbf{task} & \textbf{average}    & \textbf{Ntest} & \textbf{Npos}  & \textbf{Ctest} & \textbf{Cpos}  & \textbf{Test avg} & \textbf{Pos avg} & \textbf{Drop} & \textbf{Pos Drop} \\ \hline
PC       & ELMo 150            & 81.95          & 72.15          & 73.25          & 49             & 77.6              & 60.58            & 10.55         & 31.9              \\ \hline
PC       & DistilBERT 150      & 80.95          & 70.7           & 72.8           & 48.6           & 76.88             & 59.65            & 10            & 31.25             \\ \hline
PC       & ELMo                & 82.17          & 72.6           & 73.43          & 49.37          & 77.8              & 60.98            & 10.57         & 31.87             \\ \hline
PC       & DistilBERT (common) & 81.97 & 72.37 & 74.4 & 51.57 & 78.18     & 61.97   & 9.2  & 28.87    \\ \hline
PC       & DistilBERT (all) & 	\textbf{82.38} & \textbf{73.13} & \textbf{75} & \textbf{52.78} & \textbf{78.69}     & \textbf{62.95}   & \textbf{8.93}  & \textbf{27.95}    \\ \hline
\end{tabular}		
\end{adjustbox}
\label{table:avg-results-pc}
\end{table}

\begin{table}[H]
\centering
\caption[Sentiment - average scores.]{SA - Average results of ELMo and DistilBERT.}
\vspace{1em}
\begin{tabular}{|l|l|l|l|l|l|}
\hline
\textbf{task}      & \textbf{average}      & \textbf{Ntest} & \textbf{Ctest} & \textbf{Test avg} & \textbf{Drop}  \\ \hline
SA & ELMo       & 77.85 & 66.75 & 72.3     & 14.25 \\ \hline
SA & DistilBERT & \textbf{79.5}  & \textbf{68.8}  & \textbf{74.15}    & \textbf{13.6} \\ \hline
\end{tabular}
\label{table:avg-results-sa}
\end{table}

\subsection{Training time and model size}
Training time, inference time, and model size should be within a reasonable range especially in real-life scenarios. But in research studies as well it becomes important. For example, the whole pipeline from hyper-parameter tuning to inference would be re-iterated when a new data set arrives. In this section, DistilBERT and ELMo are compared in terms of the training and inference time, as well as the model size. 

Training times and model sizes are compared by averaging all model configurations common to ELMo and DistilBERT. Training and inference time are summed up to a single number. According to Table \ref{table:time-size}, DistilBERT is 30\% smaller and 83\% faster than ELMo on the average of both tasks. In terms of classifier size (excluding embeddings) DistilBERT is 13\% smaller than ELMo. On the other hand, MNB and LSVM are far more efficient than DistilBERT in size and speed by being 99\% smaller and 96\% faster.

\begin{table}[H]
\centering
\caption[Average training time and model size.]{Average training time and model size of ELMo and DistilBERT.}
\vspace{1em}
\begin{adjustbox}{width=1\textwidth}
\begin{tabular}{|l|l|l|l|l|}
\hline
\textbf{task} & \textbf{model}      & \textbf{embedding size (MB)} & \textbf{model size (MB)} & \textbf{train time (secs)} \\ \hline
PC       & MNB                 & -                            & 1.1                      & 12                         \\ \hline
PC       & ELMo (frozen)               & 358                          & 75.55                    & 1584                       \\ \hline
PC       & ELMo (fine-tuned)               & 358                          & 75.55                    & 1915                       \\ \hline
PC       & ELMo (all)               & 358                          & 75.55                    & 1690                       \\ \hline
PC       & DistilBERT (frozen) & 254                 & 65.8            & 345               \\ \hline
PC       & DistilBERT (fine-tuned) & 254                 & 65.8            & 273               \\ \hline
PC       & DistilBERT (common) & \textbf{254}                 & \textbf{65.8}            & \textbf{318}               \\ \hline
              &                     &                              &                          &                            \\ \hline
SA       & LSVM                & -                            & 0.133                    & 10                         \\ \hline
SA     & ELMo (frozen)              & 358                          & 75.55                    & 533                        \\ \hline
SA     & ELMo (fine-tuned)              & 358                          & 75.55                    & 1484                        \\ \hline
SA     & ELMo (all)              & 358                          & 75.55                    & 979                        \\ \hline
SA     & DistilBERT (frozen)         & 254                & 65.8            & 228             \\ \hline
SA     & DistilBERT (fine-tuned)         & 254                 & 65.8            & 246              \\ \hline
SA     & DistilBERT (all)         & \textbf{254}                 & \textbf{65.8}            & \textbf{237}               \\ \hline
\end{tabular}
\end{adjustbox}
\label{table:time-size}
\end{table}

\section{Statistical tests}

In the previous section, it was observed that on average DistilBERT outruns ELMo. But the gap between ELMo’s and DistilBERT’s scores is not very obvious, especially in null-context. So it is worth to see if the overall superiority of DistilBERT over ELMo is indeed significant.

In this section, randomization and McNemar's tests are realized on the best performing variants of the models according to Table \ref{table:comp-baseline-pc} and \ref{table:comp-baseline-sa}. For SA, ``ELMo + BiLSTM 60 (highest 2)'' is picked as the best ELMo variant on average test scores.

The main focus is to compare ELMo and DistilBERT to each other, and also see if traditional ML methods can compete with contextual embeddings at all. Here ELMo is compared to the baselines since the gap between ELMo and the baselines was relatively smaller in the experiments. That is, three different pairings are formed: ELMo-DistilBERT, ELMo-MNB, and ELMo-SVM.


\subsection{Randomization test}
Two-tailed and one-tailed tests are conducted under the significance level of $\alpha=0.05$. The null hypothesis is rejected when $p<0.025$ in two-tailed, $p<0.05$ in one-tailed tests. The harmonic mean of non-zero p-values is reported along with zero harmonic means, separated by a / symbol (find details in Section \ref{chapter:experimental-setup}).

First, two-tailed randomization tests are realized between the selected models in the relevant tasks. The null hypothesis is that there is no significant difference between the scores of the compared models. 

\begin{table}[H]
\centering
\caption[Protest news - two-tailed randomization test.]{PC two-tailed randomization test p-value results on the best performing model variations of ELMo, DistilBERT, and MNB.}
\vspace{1em}
\begin{adjustbox}{width=1\textwidth}
\begin{tabular}{|l|l|l|l|l|l|l|l|}
\hline
\textbf{compared models} & \textbf{task} & \textbf{Ntest} & \textbf{Ctest} & \textbf{Drop} \\ \hline
ELMo-DistilBERT          & PC       & 0.43 (0.01)    & 0.38 (0.007)  & 0.59 (0.005) \\ \hline
ELMo-MNB & PC & 0.24 (0.016) & 0.54 (0) & 0.82 (0.02) \\
\hline
\end{tabular}
\end{adjustbox}
\label{table:randtest-two-pc}
\end{table}

\begin{table}[H]
\centering
\caption[Sentiment - two-tailed randomization test.]{SA - Two-tailed randomization test p-value results on the best performing model variations of ELMo, DistilBERT, and LSVM.}
\vspace{1em}
\begin{tabular}{|l|l|l|l|l|}
\hline
\textbf{compared models} & \textbf{task} & \textbf{Ntest} & \textbf{Ctest}   & \textbf{Drop}  \\ \hline
ELMo-DistilBERT          & SA            & 0.033 & \textbf{0/0.008} & 0.06           \\ \hline
ELMo-LSVM                & SA            & 0.17           & \textbf{0}       & \textbf{0.006} \\ \hline
\end{tabular}
\label{table:randtest-two-sa}
\end{table}

\begin{table}[H]
\centering
\vspace{1em}
\begin{tabular}{|l|l|l|l|l|}
\hline
\textbf{task} & \textbf{Ntest} & \textbf{Ctest}   & \textbf{Drop}  \\ \hline
PC       & 0.43 (0.01)    & 0.38 (0.007)  & 0.59 (0.005) \\ \hline
SA            & 0.033 & \textbf{0/0.008} & 0.06           \\ \hline
\end{tabular}
\caption[Two-tailed randomization test.]{Two-tailed randomization test p-value results on the best performing model variations of ELMo and DistilBERT. Positive class significance results are also provided in parenthesis for PC task due to class imbalance.}
\label{table:randtest-two-sa}
\end{table}

As the Tables \ref{table:randtest-two-pc} illustrates, there is a significant difference between positive class PC scores of ELMo and DistilBERT. The same pattern is existent between ELMo and MNB. Though ``Drop'' p-value (0.02) is barely sufficient to reject the null hypothesis, which is understandable since ``Drop'' scores of ELMo and MNB are really not so distant (10\% and 9\%) as shown in Table \ref{table:comp-baseline-pc}. 

Table \ref{table:randtest-two-sa} strongly indicates that ELMo and DistilBERT cross-context test scores differ. So do ELMo and LSVM with very small p-values close to zero. Meanwhile, the test failed to reject the null hypothesis which claims that ELMo and LSVM perform similarly in the null context.

The next step is the one-tailed randomization test. The one-tailed test is only meaningful when the null hypothesis is rejected in the two-tailed test. Therefore, the one-tailed test is only applied to the criteria below:

In PC task:
\begin{itemize}
    \item Positive class scores in null context (``Ntest pos'').
    \item Positive class scores in cross-context (``Ctest pos'').
    \item  The performance change from null to cross-context in positive class (``Drop pos'').
\end{itemize}

In SA task:
\begin{itemize}
    \item Cross-context test scores (``Ctest'').
    \item  The performance change from null to cross-context (``Drop'').
\end{itemize}

\begin{table}[H]
\centering
\caption[Protest news - one-tailed randomization test.]{PC - One-tailed randomization test p-value results on the best performing model variations of ELMo, DistilBERT, and MNB.}
\vspace{1em}
\begin{tabular}{|l|l|l|l|l|}
\hline
\textbf{compared models} & \textbf{task} & \textbf{Ntest pos} & \textbf{Ctest pos} & \textbf{Drop pos} \\ \hline
ELMo-DistilBERT          & PC            & \textbf{0/0.01}    & \textbf{0/0.004}   & \textbf{0/0.006}  \\ \hline
ELMo-MNB                 & PC            & \textbf{0.007}     & \textbf{0/0.009}   & \textbf{0.004}    \\ \hline
\end{tabular}
\label{table:randtest-one-pc}
\end{table}

\begin{table}[H]
\centering
\caption[Sentiment - one-tailed randomization test.]{SA - One-tailed randomization test p-value results on the best performing model variations of ELMo, DistilBERT, and LSVM.}
\vspace{1em}
\begin{tabular}{|l|l|l|l|}
\hline
\textbf{compared models} & \textbf{task} & \textbf{Ctest}   & \textbf{Drop}      \\ \hline
ELMo-DistilBERT          & SA            & \textbf{0/0.017} & \textbf{0.02/0.02} \\ \hline
ELMo-LSVM                & SA            & \textbf{0/0}     & \textbf{0/0.009}   \\ \hline
\end{tabular}
\label{table:randtest-one-sa}
\end{table}

As shown in Table \ref{table:randtest-one-pc} DistilBERT outperforms ELMo in PC positive class in both null and cross-context. to cross-context than ELMo in PC positive class and both classes of SA. ELMo is more robust to cross-context than MNB in PC positive class and than LSVM in both classes of SA.

Table \ref{table:randtest-one-sa} signals for a superiority of DistilBERT in cross-context robustness over ELMo in SA task. The same observation also holds for ELMo over LSVM.

Tests in both tasks strengthen the superiority of DistilBERT over ELMo, and the superiority of ELMo over the baselines in the cross-context robustness. But the baselines can compete with ELMo when training and test data resemble each other (null context).

\subsection{McNemar's test}
McNemar's test enables to see if the models differ in their failures. The null hypothesis claims they do not. Predefined upper bound for p-value is $0.05$ to reject the null hypothesis.

As shown in the Tables \ref{table:mcn-test-pc} ELMo and DistilBERT differ in predicting positive class. Whereas, ELMo and MNB are found to be differ in null context predictions. It is curious that they are found to be closely performing on India data in Table  \ref{table:randtest-two-pc} but here McNemar test indicates that they fail in different samples. Using different statistical tests can be beneficial to see such complementary information in comparing two models.

As shown in Table \ref{table:mcn-test-sa}, ELMo and DistilBERT significantly differ from each other in positive class and cross-context predictions. ELMo differs from both baselines in cross-context but no significant difference could be detected in the null context.

\begin{table}[H]
\centering
\caption[Protest - McNemar's test.]{PC - McNemar's test p-value results on the best performing model variations of ELMo, DistilBERT, and MNB.}
\vspace{1em}
\begin{tabular}{|l|l|l|l|l|l|}
\hline
\textbf{compared models} & \textbf{task} & \textbf{Ntest} & \textbf{Ntest pos} & \textbf{Ctest}    & \textbf{Ctest pos} \\ \hline
ELMo-DistilBERT          & PC       & 0.25           & \textbf{2.30E-05}  & 0.51              & \textbf{2.80E-06}  \\ \hline
ELMo-MNB                 & PC       & \textbf{0.004}         & \textbf{0.043}               & 0.09              &    0.23    \\ \hline
\end{tabular}
\label{table:mcn-test-pc}
\end{table}

\begin{table}[H]
\centering
\caption[Sentiment - McNemar's test.]{SA - McNemar's test p-value results on the best performing model variations of ELMo, DistilBERT, and LSVM.}
\vspace{1em}
\begin{tabular}{|l|l|l|l|}
\hline
\textbf{compared models} & \textbf{task} & \textbf{Ntest} & \textbf{Ctest}    \\ \hline
ELMo-DistilBERT          & SA            & \textbf{0.029} & \textbf{3.30E-23} \\ \hline
ELMo-LSVM                & SA            & 0.16           & \textbf{9.43E-18} \\ \hline
\end{tabular}
\label{table:mcn-test-sa}
\end{table}

\section{Effect of external factors}
In this section, ELMo and DistilBERT are not directly compared to each other. Instead, the effect of various changes to the models and input is examined on both ELMo and DistilBERT, separately. In this sense, this section aims at answering the questions: 

\begin{itemize}
    \item How does fine-tuning on the target task affect ELMo’s and DistilBERT’s performance?
    \item How does adding a BiLSTM to the network influence ELMO’s and DistilBERT’s performance?
    \item How does sequence length affect the performance in the PC task?
\end{itemize}

\subsection{Effect of fine-tuning}
Word embeddings can be fine-tuned, meaning that, their weights can be updated during the learning phase to adapt it more to a specific task at hand. In this section, the effect of unfreezing the model layers is examined on the model scores.

Table \ref{table:fine-db-pc} and \ref{table:fine-db-sa} show that fine-tuning slightly degraded DistilBERT's performance in PC, and barely improved SA scores. So, fine-tuning did not have any significant effect on DistilBERT's performance in either task.

\begin{table}[H]
\centering
\caption[Protest - fine-tuning DistilBERT.]{PC - Effect of fine-tuning on DistilBERT's performance.}
\vspace{1em}
\begin{adjustbox}{width=1\textwidth}
\begin{tabular}{|l|l|l|l|l|l|l|l|l|l|}
\hline
\textbf{task}      & \textbf{model}            & \textbf{Ntest} & \textbf{Npos} & \textbf{Ctest} & \textbf{Cpos} & \textbf{Test avg} & \textbf{Pos avg} & \textbf{Drop} & \textbf{Pos Drop} \\ \hline
PC            & DistilBERT 256            & \textbf{83.8}  & \textbf{75}   & \textbf{76.8}  & \textbf{56}   & \textbf{80.3}     & \textbf{65.5}    & \textbf{8.2}  & \textbf{25.2}     \\ \hline
PC            & DistilBERT ft 256         & 83.2           & 74.6          & 76.4           & 55.4          & 79.8              & 65               & \textbf{8.2}  & 25.6              \\ \hline
PC       & DistilBERT + BiLSTM 256    & \textbf{84.2}  & \textbf{76.4} & \textbf{78.4}  & \textbf{59}   & \textbf{81.3}     & \textbf{67.7}    & \textbf{7}    & \textbf{23}       \\ \hline
PC       & DistilBERT ft + BiLSTM 256 & 84             & 76.2          & 77.2           &  57.4         & 80.6              & 66.8             & 8             & 24.8               \\ \hline
\end{tabular}
\end{adjustbox}
\label{table:fine-db-pc}
\end{table}

\begin{table}[H]
\centering
\caption[Sentiment - fine-tuning DistilBERT.]{SA - Effect of fine-tuning on DistilBERT's performance.}
\vspace{1em}
\begin{tabular}{|l|l|l|l|l|l|}
\hline
\textbf{task} & \textbf{model}   & \textbf{Ntest} & \textbf{Ctest} & \textbf{Test avg} & \textbf{Drop} \\ \hline
SA     & DistilBERT 60    & \textbf{79}    & 66.8           & 72.9              & 15.4          \\ \hline
SA     & DistilBERT ft 60 & \textbf{79}    & \textbf{68}             & \textbf{73.5}     & \textbf{14.2} \\ \hline
SA     & DistilBERT + BiLSTM 60    & \textbf{80}    & \textbf{70.2}  & \textbf{75.1}     & \textbf{12.4} \\ \hline
SA     & DistilBERT ft + BiLSTM 60 & \textbf{80}    & \textbf{70.2}  & \textbf{75.1}     & \textbf{12.4} \\ \hline
\end{tabular}
\label{table:fine-db-sa}
\end{table}

\begin{table}[H]
\centering
\caption[Protest - fine-tuning ELMO.]{PC - Effect of fine-tuning on ELMo's performance.}
\vspace{1em}
\begin{adjustbox}{width=1\textwidth}
\begin{tabular}{|l|l|l|l|l|l|l|l|l|l|}
\hline
\textbf{task} & \textbf{model} & \textbf{Ntest} & \textbf{Npos} & \textbf{Ctest} & \textbf{Cpos} & \textbf{Test avg} & \textbf{Pos avg} & \textbf{Drop} & \textbf{Pos Drop} \\ \hline
PC       & ELMo 150       & 81.4           & 70.8          & \textbf{76.6}  & \textbf{55.2} & \textbf{79}       & \textbf{63}      & \textbf{5.8}  & \textbf{21.8}     \\ \hline
PC       & ELMo ft 150    & \textbf{83}    & \textbf{74}   & 72.2           & 47            & 77.6              & 60.5             & 12.8          & 36.4              \\ \hline
PC       & ELMo BiLSTM 150 & 81.4             & 71.4          & \textbf{72.2}             & \textbf{47.2}          & 76.8               & 59.3             & \textbf{11.4} & \textbf{33.8}     \\ \hline
PC       & ELMo ft + BiLSTM 150 & \textbf{82}             & \textbf{72.4}          & 72             & 46.6          & \textbf{77}                & \textbf{59.5}             &         12.2 &      35.6     \\ \hline
\end{tabular}
\end{adjustbox}
\label{table:fine-el-pc}
\end{table}

\begin{table}[H]
\centering
\caption[Sentiment - fine-tuning ELMO.]{SA - Effect of fine-tuning on ELMos's performance.}
\vspace{1em}
\begin{tabular}{|l|l|l|l|l|l|}
\hline
\textbf{task} & \textbf{model} & \textbf{Ntest} & \textbf{Ctest} & \textbf{Test avg} & \textbf{Drop} \\ \hline
SA     & ELMo 60        & \textbf{78}    & 63.6           & 70.8              & 18.4          \\ \hline
Sa     & ELMo ft 60     & 76.2           & \textbf{69}    & \textbf{72.6}     & \textbf{9.6}  \\ \hline
SA     & ELMo + BiLSTM 60    & \textbf{79}    &  67    & \textbf{73}       & 15.2 \\ \hline
SA     & ELMo ft + BiLSTM 60 & 78.2  &   \textbf{ 67.4 }          &  72.8     & \textbf{13.8}          \\ \hline
\end{tabular}
\label{table:fine-el-sa}
\end{table}

The effect of fine-tuning is more visible on ELMo. But how it affects ELMo does not match between the tasks. As illustrated in the table \ref{table:fine-el-pc}, in PC, fine-tuning improved results in the null context (India test data) but degraded the results in the cross-context (China test data). But as Table \ref{table:fine-el-sa} illustrates, the impact is the exact opposite in SA: While scores decrease in the null context (MR test data), it increased in the cross-context (CR test data).

As a more generic observation, fine-tuning does not make the models more robust to the cross-context in PC. On the other hand, in SA, it enhances the cross-context performance. It is known that unfreezing can cause overfitting on small training data \cite{howard-ruder-2018-universal}. The occasional drop in the null context performance can be a sign of overfitting.

\subsection{Effect of BiLSTM}
As the Tables \ref{table:bilstm-db-pc} and \ref{table:bilstm-db-sa} show, BiLSTM usage persistently helped increase DistilBERT’s performance for both tasks but with rather a small improvement. As illustrated on the Table \ref{table:bilstm-db-sa} the cross-context gain is most visible when DistilBERT is frozen in SA, with the improvement in the CR test score (see the ``Ctest'' column) from $66.8$ to $70.2$ F-score.

\begin{table}[H]
\centering
\caption[Protest - BiLSTM with DistilBERT.]{PC - Effect of BiLSTM usage on DistilBERT's performance. The models are compared in pairs from top to down.}
\vspace{1em}
\begin{adjustbox}{width=1\textwidth}
\begin{tabular}{|l|l|l|l|l|l|l|l|l|l|}
\hline
\textbf{task} & \textbf{model}             & \textbf{Ntest} & \textbf{Npos} & \textbf{Ctest} & \textbf{Cpos} & \textbf{Test avg} & \textbf{Pos avg} & \textbf{Drop} & \textbf{Pos Drop} \\ \hline
PC       & DistilBERT 256             & 83.8           & 75            & 76.8           & 56            & 80.3              & 65.5             & 8.2           & 25.2              \\ \hline
PC       & DistilBERT + BiLSTM 256    & \textbf{84.2}  & \textbf{76.4} & \textbf{78.4}  & \textbf{59}   & \textbf{81.3}     & \textbf{67.7}    & \textbf{7}    & \textbf{23}       \\ \hline
PC       & DistilBERT ft 256          & 83.2           & 74.6          & 76.4           & 55.4          & 79.8              & 65               & 8.2           & 25.6              \\ \hline
PC       & DistilBERT ft + BiLSTM 256 & \textbf{84}    & \textbf{76.2} & \textbf{77.2}  & \textbf{57.4} & \textbf{80.6}     & \textbf{66.8}    & \textbf{8}    & \textbf{24.8}     \\ \hline
\end{tabular}
\end{adjustbox}
\label{table:bilstm-db-pc}
\end{table}

\begin{table}[H]
\centering
\caption[Sentiment - BiLSTM with DistilBERT.]{SA - Effect of BiLSTM usage on DistilBERT's performance. The models are compared in pairs from top to down.}
\vspace{1em}
\begin{tabular}{|l|l|l|l|l|l|}
\hline
\textbf{task} & \textbf{model}            & \textbf{Ntest} & \textbf{Ctest} & \textbf{Test avg} & \textbf{Drop} \\ \hline
SA            & DistilBERT 60             & 79             & 66.8           & 72.9              & 15.4          \\ \hline
SA            & DistilBERT + BiLSTM 60    & \textbf{80}    & \textbf{70.2}  & \textbf{75.1}     & \textbf{12.4} \\ \hline
SA            & DistilBERT ft 60          & 79             & 68             & 73.5              & 14.2          \\ \hline
SA            & DistilBERT ft + BiLSTM 60 & \textbf{80}    & \textbf{70.2}  & \textbf{75.1}     & \textbf{12.4} \\ \hline
\end{tabular}
\label{table:bilstm-db-sa}
\end{table}

The effect of BiLSTM is not as persistent on ELMo as it was on DistilBERT. BiLSTM more or less always improved the DistilBERT performance, but there is no such a pattern for ELMo. As illustrated in Table \ref{table:bilstm-el-pc}, BiLSTM usage affects ELMo badly in PC, especially in the cross-context performance of frozen ELMo with drop from $52.6$ to $47.6$ F-score. Meanwhile, as Table \ref{table:bilstm-db-sa} shows, in SA BiLSTM improves cross-context robustness of frozen ELMo, but affects the fine-tuned ELMo negatively.

ELMo's inner architecture resembles a BiLSTM, whereas DistilBERT's doesn't. That could be one reason that ELMo cannot benefit from additional LSTM contextualization as much as DistilBERT. 

\begin{table}[H]
\centering
\caption[Protest - BiLSTM with ELMo.]{PC - Effect of BiLSTM usage on ELMO's performance. The models are compared in pairs from top to down.}
\vspace{1em}
\begin{adjustbox}{width=1\textwidth}
\begin{tabular}{|l|l|l|l|l|l|l|l|l|l|}
\hline
\textbf{task} & \textbf{model}       & \textbf{Ntest} & \textbf{Npos} & \textbf{Ctest} & \textbf{Cpos} & \textbf{Test avg} & \textbf{Pos avg} & \textbf{Drop} & \textbf{Pos Drop} \\ \hline
PC       & ELMo 256             & \textbf{83.6}  & \textbf{74.8} & \textbf{75.2}  & \textbf{52.6} & \textbf{79.4}     & \textbf{63.7}    & \textbf{10}   & \textbf{29.6}     \\ \hline
PC       & ELMo + BiLSTM 256    & 81.6           & 72.2          & 72.4           & 47.6          & 77                & 59.9             & 11.2          & 34                \\ \hline
PC       & ELMo ft 150          & \textbf{83}    & \textbf{74}   & \textbf{72.2}  & \textbf{47}   & \textbf{77.6}     & \textbf{60.5}    & 12.8          & 36.4              \\ \hline
PC       & ELMo ft + BiLSTM 150 & 82             & 72.4          & 72             & 46.6          & 77                & 59.5             & \textbf{12.2} & \textbf{35.6}     \\ \hline
\end{tabular}
\end{adjustbox}
\label{table:bilstm-el-pc}
\end{table}

\begin{table}[H]
\centering
\caption[Sentiment - BiLSTM with ELMo.]{SA - Effect of BiLSTM usage on ELMO's performance. The models are compared in pairs from top to down.}
\vspace{1em}
\begin{tabular}{|l|l|l|l|l|l|}
\hline
\textbf{task} & \textbf{model}      & \textbf{Ntest} & \textbf{Ctest} & \textbf{Test avg} & \textbf{Drop} \\ \hline
SA            & ELMo 60             & 78             & 63.6           & 70.8              & 18.4          \\ \hline
SA            & ELMo + BiLSTM 60    & \textbf{79}    & \textbf{67}    & \textbf{73}       & \textbf{15.2} \\ \hline
SA            & ELMo ft 60          & 76.2           & \textbf{69}    & 72.6              & \textbf{9.6}  \\ \hline
SA            & ELMo ft + BiLSTM 60 & \textbf{78.2}  & 67.4           & \textbf{72.8}     & 13.8          \\ \hline
\end{tabular}
\label{table:bilstm-el-sa}
\end{table}

\subsection{Effect of sequence length}
Sequence length is one of the parameters that NLP practitioners should sometimes restrict to a certain level due to limited device memory. In this section, the effect of the sequence length is observed in the PC task. 
When ELMo is fine-tuned, it cannot handle long sequences (256) in a single GPU. That is why in PC, the input size is decreased down to 150. But in SA, this was not needed thanks to short input length (20 on average). 

In PC, as illustrated in Table \ref{table:seqlen-db}, long sequence persistently enhances DistilBERT's performance. Longer context enabled to learn better both task-specific as well as transferable features.

\begin{table}[H]
\centering
\caption[Sequence length effect on DistilBERT.]{Effect of maximum sequence length (token-wise) on DistilBERT's performance. The models are compared in pairs from top to down.}
\vspace{1em}
\begin{adjustbox}{width=1\textwidth}
\begin{tabular}{|l|l|l|l|l|l|l|l|l|l|}
\hline
\textbf{task} & \textbf{model}             & \textbf{Ntest} & \textbf{Npos} & \textbf{Ctest} & \textbf{Cpos} & \textbf{Test avg} & \textbf{Pos avg} & \textbf{Drop} & \textbf{Pos Drop} \\ \hline
PC       & DistilBERT 150             & 80.4           & 70.2          & 75.4           & 53            & 77.9              & 61.6             & \textbf{6.2}  & \textbf{24.8}     \\ \hline
PC       & DistilBERT 256             & \textbf{83.8}  & \textbf{75}   & \textbf{76.8}  & \textbf{56}   & \textbf{80.3}     & \textbf{65.5}    & 8.2           & 25.2              \\ \hline
PC       & DistilBERT ft 150          & 80             & 69            & 71             & 45.6          & 75.5              & 57.3             & 11            & 33.8              \\ \hline
PC       & DistilBERT ft 256          & \textbf{83.2}  & \textbf{74.6} & \textbf{76.4}  & \textbf{55.4} & \textbf{79.8}     & \textbf{65}      & \textbf{8.2}  & \textbf{25.6}     \\ \hline
PC       & DistilBERT + BiLSTM 150    & 81.6           & 71.8          & 72.6           & 48.2          & 77.1              & 60               & 11            & 32.8              \\ \hline
PC       & DistilBERT + BiLSTM 256    & \textbf{84.2}  & \textbf{76.4} & \textbf{78.4}  & \textbf{59}   & \textbf{81.3}     & \textbf{67.7}    & \textbf{7}    & \textbf{23}       \\ \hline
PC       & DistilBERT ft + BiLSTM 150 & 81.8           & 71.8          & 72.2           & 47.6          & 77                & 59.7             & 11.8          & 33.6              \\ \hline
PC       & DistilBERT ft + BiLSTM 256 & \textbf{84}    & \textbf{76.2} & \textbf{77.2}  & \textbf{57.4} & \textbf{80.6}     & \textbf{66.8}    & \textbf{8}    & \textbf{24.8}     \\ \hline
\end{tabular}
\end{adjustbox}
\label{table:seqlen-db}
\end{table}

ELMo cannot always benefit from long sequences as opposed to DistilBERT. As shown in the ``Drop'' column of Table \ref{table:seqlen-el}, long sequences have occasionally a negative effect in the cross-context performance. It is known that a drawback of LSTM is that it is not very successful in learning long-term dependencies in a sequence due to the vanishing gradient phenomena \cite{howard-ruder-2018-universal}. On the other hand, transformer-based models can handle it with the attention-based technique \cite{transformer-NIPS2017_7181}.

\begin{table}[H]
\centering
\caption[Sequence length effect on ELMo.]{Effect of maximum sequence length (token-wise) on ELMo's performance. The models are compared in pairs from top to down.}
\vspace{1em}
\begin{adjustbox}{width=1\textwidth}
\begin{tabular}{|l|l|l|l|l|l|l|l|l|l|}
\hline
\textbf{task} & \textbf{model}    & \textbf{Ntest} & \textbf{Npos} & \textbf{Ctest} & \textbf{Cpos} & \textbf{Test avg} & \textbf{Pos avg} & \textbf{Drop} & \textbf{Pos Drop} \\ \hline
PC       & ELMo 150          & 81.4           & 70.8          & \textbf{76.6}  & \textbf{55.2} & 79                & 63               & \textbf{5.8}  & \textbf{21.8}     \\ \hline
PC       & ELMo 256          & \textbf{83.6}  & \textbf{74.8} & 75.2           & 52.6          & \textbf{79.4}     & \textbf{63.7}    & 10            & 29.6              \\ \hline
PC       & ELMo + BiLSTM 150 & 81.4           & 71.4          & 72.2           & 47.2          & 76.8              & 59.3             & 11.4          & 33.8              \\ \hline
PC       & ELMo + BiLSTM 256 & \textbf{81.6}  & \textbf{72.2} & \textbf{72.4}  & \textbf{47.6} & \textbf{77}       & \textbf{59.9}    & \textbf{11.2} & \textbf{34}       \\ \hline
\end{tabular}
\end{adjustbox}
\label{table:seqlen-el}
\end{table}
\section{Comparing BERT and DistilBERT}
DistilBERT is preferred over BERT in this study due to memory limits of single GPU. But in this section BERT results with limited sequence length of 150 tokens is reported to be able to also compare BERT (\texttt{bert-base-uncased}) \cite{devlin-etal-2019-bert}  and DistilBERT in cross-context robustness. 

As Table \ref{table:pc-bertdbert} displays, BERT and DistilBERT perform very closely on PC on all tracks. But DistilBERT seems to outrun BERT (75.4 vs. 69 F-score) on cross-context robustness when both are frozen and paired with a FFNN classifier. 

In Table \ref{table:sa-bertdbert} it is shown that BERT outperformed DistilBERT on SA especially in cross-context robustness. 

\begin{table}[H]
\centering
\caption[Protest - BERT vs. DistilBERT.]{PC results on BERT and DistilBERT.}
\vspace{1em}
\begin{adjustbox}{width=1\textwidth}
\begin{tabular}{|l|l|l|l|l|l|l|l|l|l|}
\hline
\textbf{task} & \textbf{model}             & \textbf{Ntest} & \textbf{Npos} & \textbf{Ctest} & \textbf{Cpos} & \textbf{Test avg} & \textbf{Pos avg} & \textbf{Drop} & \textbf{Pos Drop} \\ \hline
PC            & BERT 150                   & 80.8           & 70            & 69             & 41            & 74.9              & 55.5             & 14.8          & 41.4              \\ \hline
PC            & BERT ft 150                & 81             & 70.8          & 70.6           & 44.8          & 75.8              & 57.8             & 12.8          & 36.8              \\ \hline
PC            & BERT + BiLSTM 150          & 81.8           & 72            & 72.6           & 48            & 77.2              & 60               & 11            & 33.6              \\ \hline
PC            & BERT ft + BiLSTM 150       & 81.8           & 72            & 72.6           & 48            & 77.2              & 60               & 11            & 33.6              \\ \hline
PC            & BERT avg                   & \textbf{81.35} & \textbf{71.2} & 71.2           & 45.45         & 76.28             & 58.33            & 12.4          & 36.35             \\ \hline
PC            & DistilBERT 150             & 80.4           & 70.2          & 75.4           & 53            & 77.9              & 61.6             & 6.2           & 24.8              \\ \hline
PC            & DistilBERT ft 150          & 80             & 69            & 71             & 45.6          & 75.5              & 57.3             & 11            & 33.8              \\ \hline
PC            & DistilBERT + BiLSTM 150    & 81.6           & 71.8          & 72.6           & 48.2          & 77.1              & 60               & 11            & 32.8              \\ \hline
PC            & DistilBERT ft + BiLSTM 150 & 81.8           & 71.8          & 72.2           & 47.6          & 77                & 59.7             & 11.8          & 33.6              \\ \hline
PC            & DistilBERT avg             & 80.95          & 70.7          & \textbf{72.8}  & \textbf{48.6} & \textbf{76.88}    & \textbf{59.65}   & \textbf{10}   & \textbf{31.25}    \\ \hline
\end{tabular}
\end{adjustbox}
\label{table:pc-bertdbert}
\end{table}

\begin{table}[H]
\centering
\caption[Sentiment - BERT vs. DistilBERT.]{SA results on BERT and DistilBERT.}
\vspace{1em}
\begin{tabular}{|l|l|l|l|l|l|}
\hline
\textbf{task} & \textbf{model}            & \textbf{Ntest} & \textbf{Ctest} & \textbf{Test avg} & \textbf{Drop} \\ \hline
SA            & BERT 60                   & 80.6           & 75             & 77.8              & 7             \\ \hline
SA            & BERT ft 60                & 80.6           & 75             & 77.8              & 7             \\ \hline
SA            & BERT + BiLSTM 60          & 82             & 71.8           & 76.9              & 12.6          \\ \hline
SA            & BERT ft + BiLSTM 60       & 82.6           & 75.6           & 79.1              & 8.6           \\ \hline
SA            & BERT avg                  & \textbf{81.45} & \textbf{74.35} & \textbf{77.9}     & \textbf{8.8}  \\ \hline
SA            & DistilBERT 60             & 79             & 66.8           & 72.9              & 15.4          \\ \hline
SA            & DistilBERT ft 60          & 79             & 68             & 73.5              & 14.2          \\ \hline
SA            & DistilBERT + BiLSTM 60    & 80             & 70.2           & 75.1              & 12.4          \\ \hline
SA            & DistilBERT ft + BiLSTM 60 & 80             & 70.2           & 75.1              & 12.4          \\ \hline
SA            & DistilBERT avg            & 79.5           & 68.8           & 74.15             & 13.6          \\ \hline
\end{tabular}
\label{table:sa-bertdbert}
\end{table}

No randomization test is conducted on PC task between BERT and DistilBERT since the results are pretty close to each other on that task. But in SA, best performing variations of the two models (DistilBERT ft + BiLSTM 60 and BERT ft + BiLSTM 60) are compared with randomization test, since here BERT seems to outperform DistilBERT with a larger margin especially in cross-context robustness. As seen in Table \ref{table:randtest-b-db}, BERT performs significantly better in all tracks, meaning, in null, cross-context, and the ``Drop'' performance. 

\begin{table}[H]
\centering
\caption[Sentiment - BERT vs. DistilBERT - randomization test.]{SA - Two-tailed and one-tailed randomization test p-value results on the best performing model variations of BERT and DistilBERT.}
\vspace{1em}
\begin{tabular}{|l|l|l|l|l|l|}
\hline
\textbf{compared models} & \textbf{test type} & \textbf{task} & \textbf{Ntest} & \textbf{Ctest} & \textbf{Drop} \\ \hline
DistilBERT-BERT          & two-tailed         & SA            & 0/0.008        & 0/0.03         & 0/0.009       \\ \hline
DistilBERT-BERT          & one-tailed         & SA            & 0/0.003        & 0/0.018        &    0/0.004           \\ \hline
\end{tabular}
\label{table:randtest-b-db}
\end{table}

\section{New state-of-the-art in CLEF-2019 Lab ProtestNews}
If time and resources allow, hyper-parameter tuning can effectively improve results. In this study, combining contextual embeddings with standard shallow neural networks (FFNN and BiLSTM) and applying hyper-parameter tuning helped outrun the prior results in the CLEF-2019 Lab ProtestNews in cross-context while getting comparable results in null context. As shown in Table \ref{table:shared-task-comp}, F-score in China test set increased from 65 to 76.8 F-score; ``Drop'' is diminished from 22\% to 8.2\%.

\begin{table}[H]
\centering
\caption[Comparison with CLEF-2019 Lab ProtestNews results.]{Comparison with CLEF-2019 Lab ProtestNews results. The prior state-of-the-art is exceeded in cross-context.}
\vspace{1em}
\begin{tabular}{|l|l|l|l|}
\hline
\textbf{Ntest}             & \textbf{Ntest} & \textbf{Ctset} & \textbf{Drop}  \\ \hline
Radford et al. \cite{radford-clef-levelup}           & 83             & 65            & 22         \\ \hline
DistilBERT 256 & \textbf{83.8}    & \textbf{76.8} & \textbf{8.2} \\ \hline
\end{tabular}
\label{table:shared-task-comp}
\end{table}

\chapter{DISCUSSION}
\label{chapter:discussion}
DistilBERT is better at utilizing longer sequences than ELMo. Fine-tuned ELMo cannot handle as many tokens as DistilBERT can due to excessive RAM usage. This deteriorates ELMo's performance, especially in the cross-context. Moreover, fine-tuning causes training ELMo to last 1.5X longer, while the effect is negligible in DistilBERT. 

One possible reason for DistilBERT’s overall superiority in both binary classification tasks over ELMo might be that transformer-based models can catch long-term dependencies in a sequence input better than LSTM-based models.

One may notice that there is almost always a discrepancy between overall data set statistics and positive class (protest) statistics in PC. For example, DistilBERT almost always outperforms ELMo in positive class statistics while fails to do so in general scores such as Ctest and Drop. This pattern is absent in SA statistics. This phenomenon can be associated with the class imbalance in the protest news data set. At this point, one wonders whether the results would be in favor of DistilBERT if it is not for class-imbalance. This question is left unanswered within the scope of this work.

One remark should be made that since statistical tests are not transitive, no definite conclusions can be drawn on comparing DistilBERT and the baselines. But as mentioned earlier, this was not among the main objectives of the study.

Theoretically, the data sets are large enough to realize statistical tests,  but more powerful conclusions could be drawn with larger data. 

Adding a BiLSTM to the classifier network, or fine-tuning the pretrained embeddings on the training data of the target task are two obvious steps for better results in general. But one must be aware that adding more complexity to a neural network or fine-tuning does not necessarily improve results. The real effect of all of these attempts is eventually strictly bounded by the training data and the task itself. The imbalance between data classes, the noise within the data, the complexity of the classification task, all of these can make a presumably simple binary classification problem a rather complex one. 

One interesting observation is that null context performance and cross-context performance do not necessarily grow together. For some specific configurations, when DistilBERT outran ELMo in the null context, ELMo happened to outperform DistilBERT in the cross-context or vice versa. Similarly, fine-tuning could improve null context performance but cause a drop in the cross-context performance. Even usage of longer context can cause such an effect. These observations indicate that it is important to check the robustness of a model on multiple dimensions to understand true generalization power.

It should be emphasized that the limitations of the experimental setup and the scope must always be noted when the observations of this study is concerned. All conclusions are valid only under the specific experimental setup of this study, comprising of the aforementioned binary classification tasks and the data sets. The results would be completely different even if the models were pretrained with any other corpora out there. So it must be underlined that the comparison results are special to the model-unlabeled data combinations (ELMo combined with One Billion Word Benchmark, DistilBERT combined with English Wikipedia and Toronto BookCorpus).

In this study, ELMo and DistilBERT are compared on their fine-tuning performance on two binary text classification tasks. The main focus was to see how much can these models be benefited in a practical way without any modification to the pretraining outputs. But the models were actually pretrained on entirely different corpora (ELMo on One Billion Words Benchmark \cite{onebillion}, DistilBERT on English Wikipedia and Toronto BookCorpus \cite{bookcorpus}). If the models were also pretrained from scratch on the same corpus, it would be ensured that they utilize the same knowledge to learn the context. And this would enable a fairer comparison.

Recently, it was shown that ELMo and BERT make no significant difference in semantic analysis \cite{peters-etal-2019-tune}. Here it is observed that although they are close-by in the null context, DistilBERT is more robust than ELMo in the cross-context in text classification. 

The findings of this study are in line with prior work. The fairly comparable scores of ELMo and the traditional baselines in the null context supports the observation of \cite{tenney-47786} that is, when it comes to contextual embeddings, there is only a small improvement in learning semantics over traditional ML methods. DistilBERT is on par with or exceeding ELMo on a binary text classification task \cite{peters-etal-2019-tune}. DistilBERT, as a transformer-based model, is better in capturing long-term dependencies in an input sequence \cite{transformer-NIPS2017_7181}. DistilBERT is lighter than ELMo and has a shorter training time \cite{sanh2019distilbert}.

Here it should be noted that the experimental settings of the previous work and this study are completely different in terms of compared models, data sets, classifiers, training pipeline in general. So a genuine comparison of the observations is simply not possible. This is the general problem faced in research of evaluation: the NLP field is far from reaching generic conclusions about comparing various models. But anyways, the conductor of this thesis study humbly considers that here recalling the findings of previous studies might be more beneficial than ignoring them altogether. So the remarks of the previous related work are associated with the remarks of this thesis study, but by completely being aware of the problematic situation.

\chapter{CONCLUSION}
\label{chapter:conclusion}
In this study, two state-of-the-art contextual language representations, ELMo and DistilBERT, are compared in an extrinsic manner, through two different text classification tasks, namely, protest classification and sentiment analysis. In both tasks, two different test data sets are used: one as a subsection of the corpus that the training data comes from, the other is a test set coming from a distinct country and a domain, in news classification and sentiment analysis, respectively. 

The main motive is to contribute to the evaluation space of these two competing contextual language representations by comparing them on the robustness to unseen test data coming from unfamiliar context, which is one of the main indicators that tell about generalization capacity. With this respect, the experiments are designed around a cross-context setup. The comparison is done focusing on three different criteria: F-score on the test data (null context), F-score on the cross-context test data, and the amount of F-score degradation in percentages when shifted from null to cross-context.

The models are compared in several environments. First, they are simply used as fixed word vectors when paired with a simple shallow FFNN. Second, they are fine-tuned on the training data. Third, they are paired with a BiLSTM. Lastly, they are both fine-tuned and fed into a 2-layer BiLSTM. MNB and LSVM results are also reported as traditional ML model results. For each of these variations, for each model, hyper-parameter tuning is applied. Each variation is trained with the tuned hyper-parameters with 5 different seeds to get the average performance. After the experiments with these variations, two statistical tests, randomization and McNemar’s tests, are performed, to draw more confident conclusions. 

This study methodologically differs from the previous work on evaluating the generalizability by using a new socio-political and local news data set other than vigorously utilized data sets. Second, the comparison is consistently performed under identical conditions and on a cross-context data, without any domain adaptation. 

Overall, DistilBERT is found to generalize better than ELMo on the cross-context, in addition to being  30\% smaller in embedding size and  83\% faster in training time. No significant difference could be detected between ELMo and DistilBERT in the null context. The baselines are outran by both models in the cross-context robustness. But baselines could occasionally get comparable results with ELMo in the null context. Also, they are very economic with 99\% smaller size and 96\% faster training and testing time when compared to DistilBERT.

As a result, when the transfer power of a model is a priority, it is worth to prefer contextual neural models over traditional ML methods despite much longer training times and memory overhead. On the other hand, traditional ML methods might still be preferred as low-cost options when there is no anticipated discrepancy between training and test data.
 


\chapter{FUTURE WORK}
\label{chapter:future-work}

This study can be enhanced further in various dimensions. As to the implementation dimension, currently, ELMo and DistilBERT tokenize data differently: the former uses Spacy tokenizer, the latter applies WordPiece tokenization. A fairer comparison could be performed by tokenizing the input in same way. One technical discrepancy occurs in how the models’ hidden layers are combined to get a token representation. ELMo’s hidden LSTM layers are weighted averaged (scalar-mix method), which is the default combination technique of ELMo. On the other hand, DistilBERT’s last four hidden layers are simply averaged, which is a suggested usage of BERT, the model DistilBERT is derived from, for text classification tasks. Also, for a more suitable BiLSTM implementation, BiLSTM hidden layer dimension could be set equal to the embedding dimension, instead of fixing to some arbitrary commonly used value of 512. 

Currently, SA is applied to a custom split of the MR data set. K-fold cross-validation could be applied to the MR data. This way, the results would be comparable to the prior work. At least, the results could be averaged over $n$ such custom splits to get a more realistic idea on score distributions.

ELMo and DistilBERT could also be paired with the baseline models MNB and LSVM to see the real benefit of contextualization of word vectors over traditional representations such as tf-idf and Bag of Words; and also to see performance of ELMo and DistilBERT without benefiting neural network classifiers such as FFNN and BiLSTM. 

In this study, ELMo and DistilBERT are compared on their fine-tuning performance on two binary text classification tasks. The main focus was to see how much can these models be benefited in a practical way without any modification to the pretraining outputs. But the models were actually pretrained on entirely different corpora (ELMo on One Billion Words Benchmark \cite{onebillion}, DistilBERT on English Wikipedia and Toronto BookCorpus \cite{bookcorpus}). If the models were also pretrained from scratch on the same corpus, it would be ensured that they utilize the same knowledge to learn the context. This would enable a fairer comparison.

To better see the effect of contextualization, a non-contextual word representation such as GloVe could be added to the comparisons. Also, a class-balanced document-level data set can be added to the analysis to see the vulnerability of models to class imbalance. Addition of various binary classification tasks on various data sets can help gain better understanding of the model performances.

Understanding the data has often been an underestimated part of the whole picture. However, analyzing the data beforehand could help take more realistic decisions in the succeeding stages. It allows discriminating between the effect of data or task complexity, and the model competency during error analysis. Also, suitable preprocessing steps would improve the results and alleviate the noise in the data.

An extensive error analysis could be done to explain what exactly is learned by the models. This could be done in two main ways: by conventional error analysis methods on predicted development data or focusing on interpreting the black box of the deep neural networks. On the conventional error analysis track, true positive and false negative intersections could be examined manually. On the more innovative track, the hidden state weights of the models could be visualized to understand what the models attend.

Generalization in NLP attracts much attention thanks to two main factors: 1) Most language representations can be pretrained in an unsupervised manner on a large unlabeled data. 2) Once they are pretrained, they can be used as a sane initialization point for many NLP tasks. This way task-specific models can be trained on a little labelled data for a short time. So, currently task-specific training is still bound to labeled data. But actually task-specific training can very much profit from unsupervised learning: 1) Data annotation phase would be shortened and deployment of NLP systems could be faster. 2) With diverse unlabelled data, models could be more robust to different domains. For example, in this study the classifiers could adapt to many different cross-context settings more effectively and much faster. Therefore, unsupervised domain adaptation seems to be a wise direction to take \cite{han2019unsupervised}. 

Also, analysing conflicting information within the same data can help explaining the decision of the models. Confidence scores can also be incorporated into training to interpret the gray area better \cite{NIPS2019_8556}.

Diversity in data is a requirement for models to acquire transferable features of language. But as Bottou et al. \cite{bottou2019-irm} highlights, models might learn spurious correlations from a diverse data set. This phenomena hinders model from acquiring the casual knowledge. To prevent this issue, the subsets of the diverse data set that are coming from different environments can be incorporated into the training phase in a controlled manner. Bottou et al. \cite{bottou2019-irm} proposed a method to minimize the average risk across all environments. It would be curious to see the effect of such an approach in cross-context robustness. 

Currently NLP evaluation and comparison studies are realized under varying conditions defined by specific priorities and research interests of every other study, including this particular one. This prevents making proper comparisons between observations of studies, which could enable progress based on a much more confident common ground. Defining standard evaluation pipelines to be adopted within the NLP field in general can be a way to overcome this dilemma. 

Humans indeed have natural ability to map concepts to practicality. But how it is achieved remains as a dilemma introducing such questions: How the level of abstraction is determined? How is the cost function defined? How are concepts incorporated into general knowledge? How is hierarchy of concepts handled? How is any conflict between definitions handled? What about uncertainty? 

To be able to create NLP systems that are robust to the signal complexity and variety human face in real life, such questions should be tackled with. It is curious to think how all this can eventually boil down to modeling the data better. In this path, evaluating the state-of-the-art is of vital importance to find even better data representations. And doing this evaluation in a well-defined and systematic way can accelerate progress by enabling reproducibility and collaboration. 

\bibliographystyle{styles/fbe_tez_v11}
\bibliography{references}

\begin{thebibliography}{10}
\newcommand{\enquote}[1]{``#1''}
\expandafter\ifx\csname url\endcsname\relax
  \def\url#1{{\tt #1}}\fi
\expandafter\ifx\csname urlprefix\endcsname\relax\def\urlprefix{}\fi

\bibitem{10.3389/fpsyg.2017.01918}
Sturdy, C.~B. and E.~Nicoladis, \enquote{How Much of Language Acquisition Does
  Operant Conditioning Explain?}, {\em Frontiers in Psychology\/}, Vol.~8, p.
  1918, 2017,
  \urlprefix\url{https://www.frontiersin.org/article/10.3389/fpsyg.2017.01918}.

\bibitem{ettinger-generalizability-st}
Ettinger, A., S.~Rao, H.~Daum{\'e}~III and E.~M. Bender, \enquote{Towards
  Linguistically Generalizable {NLP} Systems: A Workshop and Shared Task}, {\em
  Proceedings of the First Workshop on Building Linguistically Generalizable
  {NLP} Systems\/}, pp. 1--10, Association for Computational Linguistics,
  Copenhagen, Denmark, Sep. 2017,
  \urlprefix\url{https://www.aclweb.org/anthology/W17-5401}.

\bibitem{hurriyetoglu-CLEF}
H{\"u}rriyeto{\u{g}}lu, A., E.~Y{\"o}r{\"u}k, D.~Y{\"u}ret, {\c{C}}.~Yoltar,
  B.~G{\"u}rel, F.~Duru{\c{s}}an and O.~Mutlu, \enquote{A Task Set Proposal for
  Automatic Protest Information Collection Across Multiple Countries},
  L.~Azzopardi, B.~Stein, N.~Fuhr, P.~Mayr, C.~Hauff and D.~Hiemstra (Editors),
  {\em Advances in Information Retrieval\/}, pp. 316--323, Springer
  International Publishing, Cham, 2019.

\bibitem{word2vec-NIPS2013_5021}
Mikolov, T., I.~Sutskever, K.~Chen, G.~S. Corrado and J.~Dean,
  \enquote{Distributed Representations of Words and Phrases and their
  Compositionality}, C.~J.~C. Burges, L.~Bottou, M.~Welling, Z.~Ghahramani and
  K.~Q. Weinberger (Editors), {\em Advances in Neural Information Processing
  Systems 26\/}, pp. 3111--3119, Curran Associates, Inc., 2013,
  \urlprefix\url{http://papers.nips.cc/paper/5021-distributed-representations-of-words-and-phrases-and-their-compositionality.pdf}.

\bibitem{peters-etal-2018-deep}
Peters, M., M.~Neumann, M.~Iyyer, M.~Gardner, C.~Clark, K.~Lee and
  L.~Zettlemoyer, \enquote{Deep Contextualized Word Representations}, {\em
  Proceedings of the 2018 Conference of the North {A}merican Chapter of the
  Association for Computational Linguistics: Human Language Technologies,
  Volume 1 (Long Papers)\/}, pp. 2227--2237, Association for Computational
  Linguistics, New Orleans, Louisiana, Jun. 2018,
  \urlprefix\url{https://www.aclweb.org/anthology/N18-1202}.

\bibitem{devlin-etal-2019-bert}
Devlin, J., M.-W. Chang, K.~Lee and K.~Toutanova, \enquote{{BERT}: Pre-training
  of Deep Bidirectional Transformers for Language Understanding}, {\em
  Proceedings of the 2019 Conference of the North {A}merican Chapter of the
  Association for Computational Linguistics: Human Language Technologies,
  Volume 1 (Long and Short Papers)\/}, pp. 4171--4186, Association for
  Computational Linguistics, Minneapolis, Minnesota, Jun. 2019,
  \urlprefix\url{https://www.aclweb.org/anthology/N19-1423}.

\bibitem{radford2019language}
Radford, A., J.~Wu, R.~Child, D.~Luan, D.~Amodei and I.~Sutskever, {\em
  Language Models are Unsupervised Multitask Learners\/}, Tech. rep., Open AI,
  2019.

\bibitem{sanh2019distilbert}
Sanh, V., L.~Debut, J.~Chaumond and T.~Wolf, \enquote{DistilBERT, a distilled
  version of BERT: smaller, faster, cheaper and lighter}, {\em NeurIPS $EMC^2$
  Workshop\/}, 2019.

\bibitem{grandstrand:2004}
Greenwood, A., \enquote{Computational Neuroscience: A Window to Understanding
  How the Brain Works}, {\em Science at the Frontier\/}, chap.~9, pp. 199--232,
  The National Academies Press, Washington, DC, 1992.

\bibitem{KELL2018630}
Kell, A.~J., D.~L. Yamins, E.~N. Shook, S.~V. Norman-Haignere and J.~H.
  McDermott, \enquote{A Task-Optimized Neural Network Replicates Human Auditory
  Behavior, Predicts Brain Responses, and Reveals a Cortical Processing
  Hierarchy}, {\em Neuron\/}, Vol.~98, No.~3, pp. 630 -- 644.e16, 2018,
  \urlprefix\url{http://www.sciencedirect.com/science/article/pii/S0896627318302502}.

\bibitem{Collobert:2011:NLP:1953048.2078186}
Collobert, R., J.~Weston, L.~Bottou, M.~Karlen, K.~Kavukcuoglu and P.~Kuksa,
  \enquote{Natural Language Processing (Almost) from Scratch}, {\em J. Mach.
  Learn. Res.\/}, Vol.~12, pp. 2493--2537, Nov. 2011,
  \urlprefix\url{http://dl.acm.org/citation.cfm?id=1953048.2078186}.

\bibitem{Mikolov-skip-gram-41224}
Mikolov, T., G.~Corrado, K.~Chen and J.~Dean, \enquote{Efficient Estimation of
  Word Representations in Vector Space}, pp. 1--12, 01 2013.

\bibitem{pennington-etal-2014-glove}
Pennington, J., R.~Socher and C.~Manning, \enquote{{G}love: Global Vectors for
  Word Representation}, {\em Proceedings of the 2014 Conference on Empirical
  Methods in Natural Language Processing ({EMNLP})\/}, pp. 1532--1543,
  Association for Computational Linguistics, Doha, Qatar, Oct. 2014,
  \urlprefix\url{https://www.aclweb.org/anthology/D14-1162}.

\bibitem{kiros-skipthought-NIPS2015_5950}
Kiros, R., Y.~Zhu, R.~R. Salakhutdinov, R.~Zemel, R.~Urtasun, A.~Torralba and
  S.~Fidler, \enquote{Skip-Thought Vectors}, C.~Cortes, N.~D. Lawrence, D.~D.
  Lee, M.~Sugiyama and R.~Garnett (Editors), {\em Advances in Neural
  Information Processing Systems 28\/}, pp. 3294--3302, Curran Associates,
  Inc., 2015,
  \urlprefix\url{http://papers.nips.cc/paper/5950-skip-thought-vectors.pdf}.

\bibitem{bojanowski-etal-2017-enriching}
Bojanowski, P., E.~Grave, A.~Joulin and T.~Mikolov, \enquote{Enriching Word
  Vectors with Subword Information}, {\em Transactions of the Association for
  Computational Linguistics\/}, Vol.~5, pp. 135--146, 2017,
  \urlprefix\url{https://www.aclweb.org/anthology/Q17-1010}.

\bibitem{cove-NIPS2017_7209}
McCann, B., J.~Bradbury, C.~Xiong and R.~Socher, \enquote{Learned in
  Translation: Contextualized Word Vectors}, I.~Guyon, U.~V. Luxburg,
  S.~Bengio, H.~Wallach, R.~Fergus, S.~Vishwanathan and R.~Garnett (Editors),
  {\em Advances in Neural Information Processing Systems 30\/}, pp. 6294--6305,
  Curran Associates, Inc., 2017,
  \urlprefix\url{http://papers.nips.cc/paper/7209-learned-in-translation-contextualized-word-vectors.pdf}.

\bibitem{transformer-NIPS2017_7181}
Vaswani, A., N.~Shazeer, N.~Parmar, J.~Uszkoreit, L.~Jones, A.~N. Gomez, L.~u.
  Kaiser and I.~Polosukhin, \enquote{Attention is All you Need}, I.~Guyon,
  U.~V. Luxburg, S.~Bengio, H.~Wallach, R.~Fergus, S.~Vishwanathan and
  R.~Garnett (Editors), {\em Advances in Neural Information Processing Systems
  30\/}, pp. 5998--6008, Curran Associates, Inc., 2017,
  \urlprefix\url{http://papers.nips.cc/paper/7181-attention-is-all-you-need.pdf}.

\bibitem{howard-ruder-2018-universal}
Howard, J. and S.~Ruder, \enquote{Universal Language Model Fine-tuning for Text
  Classification}, {\em Proceedings of the 56th Annual Meeting of the
  Association for Computational Linguistics (Volume 1: Long Papers)\/}, pp.
  328--339, Association for Computational Linguistics, Melbourne, Australia,
  Jul. 2018, \urlprefix\url{https://www.aclweb.org/anthology/P18-1031}.

\bibitem{DBLP:conf/iclr/MerityX0S17}
Merity, S., C.~Xiong, J.~Bradbury and R.~Socher, \enquote{Pointer Sentinel
  Mixture Models}, {\em 5th International Conference on Learning
  Representations, {ICLR} 2017, Toulon, France, April 24-26, 2017, Conference
  Track Proceedings\/}, 2017,
  \urlprefix\url{https://openreview.net/forum?id=Byj72udxe}.

\bibitem{lan2020albert}
Lan, Z., M.~Chen, S.~Goodman, K.~Gimpel, P.~Sharma and R.~Soricut,
  \enquote{{\{}ALBERT{\}}: A Lite {\{}BERT{\}} for Self-supervised Learning of
  Language Representations}, {\em International Conference on Learning
  Representations\/}, 2020,
  \urlprefix\url{https://openreview.net/forum?id=H1eA7AEtvS}.

\bibitem{liu2019roberta}
Liu, Y., M.~Ott, N.~Goyal, J.~Du, M.~Joshi, D.~Chen, O.~Levy, M.~Lewis,
  L.~Zettlemoyer and V.~Stoyanov, \enquote{RoBERTa: A Robustly Optimized BERT
  Pretraining Approach}, {\em arXiv:1907.11692\/}, 2019.

\bibitem{xlnet-NIPS2019_8812}
Yang, Z., Z.~Dai, Y.~Yang, J.~Carbonell, R.~R. Salakhutdinov and Q.~V. Le,
  \enquote{XLNet: Generalized Autoregressive Pretraining for Language
  Understanding}, H.~Wallach, H.~Larochelle, A.~Beygelzimer,
  F.~d\textquotesingle Alch\'{e}-Buc, E.~Fox and R.~Garnett (Editors), {\em
  Advances in Neural Information Processing Systems 32\/}, pp. 5754--5764,
  Curran Associates, Inc., 2019,
  \urlprefix\url{http://papers.nips.cc/paper/8812-xlnet-generalized-autoregressive-pretraining-for-language-understanding.pdf}.

\bibitem{zhang2019ernie}
Zhang, Z., X.~Han, Z.~Liu, X.~Jiang, M.~Sun and Q.~Liu, \enquote{{ERNIE}:
  Enhanced Language Representation with Informative Entities}, {\em Proceedings
  of ACL 2019\/}, 2019.

\bibitem{Beltagy2019SciBERT}
Beltagy, I., K.~Lo and A.~Cohan, \enquote{SciBERT: Pretrained Language Model
  for Scientific Text}, {\em EMNLP\/}, 2019.

\bibitem{biobert-10.1093/bioinformatics/btz682}
Lee, J., W.~Yoon, S.~Kim, D.~Kim, S.~Kim, C.~H. So and J.~Kang,
  \enquote{{BioBERT: a pre-trained biomedical language representation model for
  biomedical text mining}}, {\em Bioinformatics\/}, 09 2019,
  \urlprefix\url{https://doi.org/10.1093/bioinformatics/btz682}.

\bibitem{liu-etal-2019-linguistic}
Liu, N.~F., M.~Gardner, Y.~Belinkov, M.~E. Peters and N.~A. Smith,
  \enquote{Linguistic Knowledge and Transferability of Contextual
  Representations}, {\em Proceedings of the 2019 Conference of the North
  {A}merican Chapter of the Association for Computational Linguistics: Human
  Language Technologies, Volume 1 (Long and Short Papers)\/}, pp. 1073--1094,
  Association for Computational Linguistics, Minneapolis, Minnesota, Jun. 2019,
  \urlprefix\url{https://www.aclweb.org/anthology/N19-1112}.

\bibitem{tenney-47786}
Tenney, I., P.~Xia, B.~Chen, A.~Wang, A.~Poliak, R.~T. McCoy, N.~Kim, B.~V.
  Durme, S.~R. Bowman, D.~Das and E.~Pavlick, \enquote{What do you learn from
  context? Probing for sentence structure in contextualized word
  representations}, {\em International Conference on Learning
  Representations\/}, 2019,
  \urlprefix\url{https://openreview.net/forum?id=SJzSgnRcKX}.

\bibitem{peters-etal-2019-tune}
Peters, M.~E., S.~Ruder and N.~A. Smith, \enquote{To Tune or Not to Tune?
  Adapting Pretrained Representations to Diverse Tasks}, {\em Proceedings of
  the 4th Workshop on Representation Learning for NLP (RepL4NLP-2019)\/}, pp.
  7--14, Association for Computational Linguistics, Florence, Italy, Aug. 2019,
  \urlprefix\url{https://www.aclweb.org/anthology/W19-4302}.

\bibitem{sun-finetune-bert-10.1007/978-3-030-32381-3_16}
Sun, C., X.~Qiu, Y.~Xu and X.~Huang, \enquote{How to Fine-Tune BERT for Text
  Classification?}, M.~Sun, X.~Huang, H.~Ji, Z.~Liu and Y.~Liu (Editors), {\em
  Chinese Computational Linguistics\/}, pp. 194--206, Springer International
  Publishing, Cham, 2019.

\bibitem{han2019unsupervised}
Han, X. and J.~Eisenstein, \enquote{Unsupervised Domain Adaptation of
  Contextualized Embeddings for Sequence Labeling}, {\em arXiv:1907.11692\/},
  2019.

\bibitem{hammond-machine-coded-event}
Hammond, J. and N.~Weidmann, \enquote{Using Machine-Coded Event Data for the
  Micro-Level Study of Political Violence}, {\em Research \& Politics\/},
  Vol.~1, 07 2014.

\bibitem{wang-monitoring-societal-events}
Wang, W., R.~Kennedy, D.~Lazer and N.~Ramakrishnan, \enquote{Growing pains for
  global monitoring of societal events}, {\em Science\/}, Vol. 353, No. 6307,
  pp. 1502--1503, 2016,
  \urlprefix\url{https://science.sciencemag.org/content/353/6307/1502}.

\bibitem{clef19protest}
H{\"u}rriyeto{\u{g}}lu, A., E.~Y{\"o}r{\"u}k, D.~Y{\"u}ret, E.~Y{\"o}r{\"u}k,
  {\c{C}}.~Yoltar, B.~G{\"u}rel, F.~Duru{\c{s}}an, O.~Mutlu, A.~Akdemir,
  T.~Gessler and P.~Makarov, {\em CLEF-2019 Lab ProtestNews on Extracting
  Protests from News\/}, 2019,
  \urlprefix\url{https://emw.ku.edu.tr/clef-protestnews-2019/}, {accessed in
  December 2019}.

\bibitem{clef19}
Conference and L.~of~the Evaluation Forum~Initiative, {\em CLEF 2019\/}, 2019,
  \urlprefix\url{http://clef2019.clef-initiative.eu/}, {accessed in December
  2019}.

\bibitem{Hurriyetoglu-overview-CLEF}
H{\"u}rriyeto{\u{g}}lu, A., E.~Y{\"o}r{\"u}k, D.~Y{\"u}ret, {\c{C}}.~Yoltar,
  B.~G{\"u}rel, F.~Duru{\c{s}}an, O.~Mutlu and A.~Akdemir, \enquote{Overview of
  CLEF 2019 Lab ProtestNews: Extracting Protests from News in a Cross-Context
  Setting}, F.~Crestani, M.~Braschler, J.~Savoy, A.~Rauber, H.~M{\"u}ller,
  D.~E. Losada, G.~Heinatz~B{\"u}rki, L.~Cappellato and N.~Ferro (Editors),
  {\em Experimental IR Meets Multilinguality, Multimodality, and
  Interaction\/}, pp. 425--432, Springer International Publishing, Cham, 2019.

\bibitem{joulin-etal-2017-fasttext}
Joulin, A., E.~Grave, P.~Bojanowski and T.~Mikolov, \enquote{Bag of Tricks for
  Efficient Text Classification}, {\em Proceedings of the 15th Conference of
  the {E}uropean Chapter of the Association for Computational Linguistics:
  Volume 2, Short Papers\/}, pp. 427--431, Association for Computational
  Linguistics, Valencia, Spain, Apr. 2017,
  \urlprefix\url{https://www.aclweb.org/anthology/E17-2068}.

\bibitem{mikolov2018advances-new-fasttext}
Mikolov, T., E.~Grave, P.~Bojanowski, C.~Puhrsch and A.~Joulin,
  \enquote{Advances in Pre-Training Distributed Word Representations}, {\em
  Proceedings of the International Conference on Language Resources and
  Evaluation (LREC 2018)\/}, 2018.

\bibitem{radford-clef-levelup}
Radford, B., \enquote{Multitask Models for Supervised Protest Detection in
  Texts}, {\em In Working Notes of CLEF 2019 - Conference and Labs of the
  Evaluation Forum\/}, 07 2019.

\bibitem{asafaya-clef}
Safaya, A., \enquote{Event Sentence Detection Task Using Attention Model}, {\em
  In Working Notes of CLEF 2019 - Conference and Labs of the Evaluation
  Forum\/}, 07 2019.

\bibitem{maslennikova-clef}
Maslennikova, E., \enquote{ELMo Word Representations For News Protection}, {\em
  In Working Notes of CLEF 2019 - Conference and Labs of the Evaluation
  Forum\/}, 07 2019.

\bibitem{Zhao:2015:SHS:2832747.2832816}
Zhao, H., Z.~Lu and P.~Poupart, \enquote{Self-adaptive Hierarchical Sentence
  Model}, {\em Proceedings of the 24th International Conference on Artificial
  Intelligence\/}, IJCAI'15, pp. 4069--4076, AAAI Press, 2015,
  \urlprefix\url{http://dl.acm.org/citation.cfm?id=2832747.2832816}.

\bibitem{conneau-etal-2017-supervised}
Conneau, A., D.~Kiela, H.~Schwenk, L.~Barrault and A.~Bordes,
  \enquote{Supervised Learning of Universal Sentence Representations from
  Natural Language Inference Data}, {\em Proceedings of the 2017 Conference on
  Empirical Methods in Natural Language Processing\/}, pp. 670--680,
  Association for Computational Linguistics, Copenhagen, Denmark, Sep. 2017,
  \urlprefix\url{https://www.aclweb.org/anthology/D17-1070}.

\bibitem{bowman-etal-2015-large}
Bowman, S.~R., G.~Angeli, C.~Potts and C.~D. Manning, \enquote{A large
  annotated corpus for learning natural language inference}, {\em Proceedings
  of the 2015 Conference on Empirical Methods in Natural Language
  Processing\/}, pp. 632--642, Association for Computational Linguistics,
  Lisbon, Portugal, Sep. 2015,
  \urlprefix\url{https://www.aclweb.org/anthology/D15-1075}.

\bibitem{conneau-kiela-2018-senteval}
Conneau, A. and D.~Kiela, \enquote{{S}ent{E}val: An Evaluation Toolkit for
  Universal Sentence Representations}, {\em Proceedings of the Eleventh
  International Conference on Language Resources and Evaluation ({LREC}
  2018)\/}, European Language Resources Association (ELRA), Miyazaki, Japan,
  May 2018, \urlprefix\url{https://www.aclweb.org/anthology/L18-1269}.

\bibitem{logeswaran2018an}
Logeswaran, L. and H.~Lee, \enquote{An efficient framework for learning
  sentence representations}, {\em International Conference on Learning
  Representations\/}, 2018,
  \urlprefix\url{https://openreview.net/forum?id=rJvJXZb0W}.

\bibitem{hill-etal-2016-learning-distributed}
Hill, F., K.~Cho and A.~Korhonen, \enquote{Learning Distributed Representations
  of Sentences from Unlabelled Data}, {\em Proceedings of the 2016 Conference
  of the North {A}merican Chapter of the Association for Computational
  Linguistics: Human Language Technologies\/}, pp. 1367--1377, Association for
  Computational Linguistics, San Diego, California, Jun. 2016,
  \urlprefix\url{https://www.aclweb.org/anthology/N16-1162}.

\bibitem{pang-lee-2005-seeing}
Pang, B. and L.~Lee, \enquote{Seeing Stars: Exploiting Class Relationships for
  Sentiment Categorization with Respect to Rating Scales}, {\em Proceedings of
  the 43rd Annual Meeting of the Association for Computational Linguistics
  ({ACL}{'}05)\/}, pp. 115--124, Association for Computational Linguistics, Ann
  Arbor, Michigan, Jun. 2005,
  \urlprefix\url{https://www.aclweb.org/anthology/P05-1015}.

\bibitem{Hu:2004:MSC:1014052.1014073}
Hu, M. and B.~Liu, \enquote{Mining and Summarizing Customer Reviews}, {\em
  Proceedings of the Tenth ACM SIGKDD International Conference on Knowledge
  Discovery and Data Mining\/}, KDD '04, pp. 168--177, ACM, New York, NY, USA,
  2004, \urlprefix\url{http://doi.acm.org/10.1145/1014052.1014073}.

\bibitem{sonmez2016towards}
S{\"o}nmez, {\c{C}}., A.~{\"O}zg{\"u}r and E.~Y{\"o}r{\"u}k, \enquote{Towards
  Building a Political Protest Database to Explain Changes in the Welfare
  State}, {\em Proceedings of the 10th ACL Workshop on Language Technology for
  Cultural Heritage, Social Sciences, and Humanities (LaTeCH 2016)\/}, pp.
  106--110, 2016.

\bibitem{ace2005events}
{\em ACE (Automatic Content Extraction) English Annotation Guidelines for
  Events\/}, 5.4.3 2005.07.01 edn., 2005.

\bibitem{cameo}
University, P.~S., {\em Conflict and Mediation Event Observations Event and
  Actor Codebook\/}, 2012,
  \urlprefix\url{http://data.gdeltproject.org/documentation/CAMEO.Manual.1.1b3.pdf},
  {accessed in December 2019}.

\bibitem{pytorch}
Paszke, A., S.~Gross, S.~Chintala and G.~Chanan, {\em PyTorch\/}, 2016,
  \urlprefix\url{https://pytorch.org/}, {accessed in December 2019}.

\bibitem{onebillion}
Chelba, C., T.~Mikolov, M.~Schuster, Q.~Ge, T.~Brants, P.~Koehn and
  T.~Robinson, {\em One Billion Word Benchmark for Measuring Progress in
  Statistical Language Modeling\/}, Tech. rep., Google, 2013,
  \urlprefix\url{http://arxiv.org/abs/1312.3005}.

\bibitem{wmt11-text}
Koehn, P., {\em EMNLP 2011 Workshop on Statistical Machine Translation\/},
  2011, \urlprefix\url{http://www.statmt.org/wmt11/}, {accessed in January
  2020}.

\bibitem{bookcorpus}
Zhu, Y., R.~Kiros, R.~Zemel, R.~Salakhutdinov, R.~Urtasun, A.~Torralba and
  S.~Fidler, \enquote{Aligning Books and Movies: Towards Story-Like Visual
  Explanations by Watching Movies and Reading Books}, {\em The IEEE
  International Conference on Computer Vision (ICCV)\/}, December 2015.

\bibitem{bucila-distil}
Buciluundefined, C., R.~Caruana and A.~Niculescu-Mizil, \enquote{Model
  Compression}, {\em Proceedings of the 12th ACM SIGKDD International
  Conference on Knowledge Discovery and Data Mining\/}, KDD ’06, p.
  535–541, Association for Computing Machinery, New York, NY, USA, 2006,
  \urlprefix\url{https://doi.org/10.1145/1150402.1150464}.

\bibitem{hinton-distil}
Hinton, G., O.~Vinyals and J.~Dean, \enquote{Distilling the Knowledge in a
  Neural Network}, {\em NIPS Deep Learning and Representation Learning
  Workshop\/}, 2015, \urlprefix\url{http://arxiv.org/abs/1503.02531}.

\bibitem{kd-nervana}
Systems, N., {\em Knowledge Distillation\/}, 2019,
  \urlprefix\url{https://nervanasystems.github.io/distiller/knowledge_distillation.html},
  {accessed in January 2020}.

\bibitem{distilbert}
HuggingFace, {\em DistilBERT\/}, 2019,
  \urlprefix\url{https://github.com/huggingface/transformers}, {accessed in
  December 2019}.

\bibitem{elmo}
AllenNLP, {\em Original ELMo model\/}, 2018,
  \urlprefix\url{https://allennlp.org/elmo}, {accessed in December 2019}.

\bibitem{scikitlearn}
Cournapeau, D., {\em scikit-learn\/}, 2007,
  \urlprefix\url{https://scikit-learn.org/}, {accessed in December 2019}.

\bibitem{adam-Kingma}
Kingma, D. and J.~Ba, \enquote{Adam: A Method for Stochastic Optimization},
  {\em International Conference on Learning Representations\/}, 12 2014.

\bibitem{cs231n}
University, S., {\em CS231n, Convolutional Neural Networks for Visual
  Recognition\/}, 2019, \urlprefix\url{http://cs231n.github.io/}, {accessed in
  January 2020}.

\bibitem{spacy}
Explosion, {\em spaCy\/}, 2014, \urlprefix\url{https://spacy.io/usage/models},
  {accessed in December 2019}.

\bibitem{45610}
Wu, Y., M.~Schuster, Z.~Chen, Q.~V. Le, M.~Norouzi, W.~Macherey, M.~Krikun,
  Y.~Cao, Q.~Gao, K.~Macherey, J.~Klingner, A.~Shah, M.~Johnson, X.~Liu,
  Łukasz Kaiser, S.~Gouws, Y.~Kato, T.~Kudo, H.~Kazawa, K.~Stevens, G.~Kurian,
  N.~Patil, W.~Wang, C.~Young, J.~Smith, J.~Riesa, A.~Rudnick, O.~Vinyals,
  G.~Corrado, M.~Hughes and J.~Dean, \enquote{Google's Neural Machine
  Translation System: Bridging the Gap between Human and Machine Translation},
  {\em CoRR\/}, Vol. abs/1609.08144, 2016,
  \urlprefix\url{http://arxiv.org/abs/1609.08144}.

\bibitem{Bergstra:2011:AHO:2986459.2986743}
Bergstra, J., R.~Bardenet, Y.~Bengio and B.~K{\'e}gl, \enquote{Algorithms for
  Hyper-parameter Optimization}, {\em Proceedings of the 24th International
  Conference on Neural Information Processing Systems\/}, NIPS'11, pp.
  2546--2554, Curran Associates Inc., USA, 2011,
  \urlprefix\url{http://dl.acm.org/citation.cfm?id=2986459.2986743}.

\bibitem{Bergstra:2013:MSM:3042817.3042832}
Bergstra, J., D.~Yamins and D.~D. Cox, \enquote{Making a Science of Model
  Search: Hyperparameter Optimization in Hundreds of Dimensions for Vision
  Architectures}, {\em Proceedings of the 30th International Conference on
  International Conference on Machine Learning - Volume 28\/}, ICML'13, pp.
  I--115--I--123, JMLR.org, 2013,
  \urlprefix\url{http://dl.acm.org/citation.cfm?id=3042817.3042832}.

\bibitem{Yeh-randtest}
Yeh, A., \enquote{More Accurate Tests for the Statistical Significance of
  Result Differences}, {\em Proceedings of the 18th Conference on Computational
  Linguistics - Volume 2\/}, COLING '00, pp. 947--953, Association for
  Computational Linguistics, Stroudsburg, PA, USA, 2000,
  \urlprefix\url{https://doi.org/10.3115/992730.992783}.

\bibitem{Dietterich:1998:AST:303222.303237}
Dietterich, T.~G., \enquote{Approximate Statistical Tests for Comparing
  Supervised Classification Learning Algorithms}, {\em Neural Comput.\/},
  Vol.~10, No.~7, pp. 1895--1923, Oct. 1998,
  \urlprefix\url{http://dx.doi.org/10.1162/089976698300017197}.

\bibitem{NIPS2019_8556}
Corbi\`{e}re, C., N.~THOME, A.~Bar-Hen, M.~Cord and P.~P\'{e}rez,
  \enquote{Addressing Failure Prediction by Learning Model Confidence},
  H.~Wallach, H.~Larochelle, A.~Beygelzimer, F.~d\textquotesingle
  Alch\'{e}-Buc, E.~Fox and R.~Garnett (Editors), {\em Advances in Neural
  Information Processing Systems 32\/}, pp. 2898--2909, Curran Associates,
  Inc., 2019,
  \urlprefix\url{http://papers.nips.cc/paper/8556-addressing-failure-prediction-by-learning-model-confidence.pdf}.

\bibitem{bottou2019-irm}
Arjovsky, M., L.~Bottou, I.~Gulrajani and D.~Lopez-Paz, \enquote{Invariant Risk
  Minimization}, {\em arXiv:1907.02893v2\/}, 2019.

\end{thebibliography}

\appendix

\chapter{Detailed PC and SA results}
\label{chapter:det-res}

\begin{table}[H]
\centering
\caption[Protest - detailed results.]{PC results including standard deviations.}
\vspace{1em}
\begin{adjustbox}{width=1\textwidth}
\begin{tabular}{|l|l|l|l|l|l|l|l|l|l|l|l|l|l|l|l|l|}
\hline
\textbf{model}             & \textbf{Ntest} & \textbf{Ntest stdev} & \textbf{Npos} & \textbf{Npos stdev} & \textbf{Ctest} & \textbf{Ctest stdev} & \textbf{Cpos} & \textbf{Cpos stdev} & \textbf{Test avg} & \textbf{Test avg stdev} & \textbf{Pos avg} & \textbf{Pos avg stdev} & \textbf{Drop} & \textbf{Drop stdev} & \textbf{Pos Drop} & \textbf{Pos Drop stdev} \\ \hline
ELMo 256                   & 83.6           & 0.55                 & 74.8          & 0.84                & 75.2           & 1.3                  & 52.6          & 2.7                 & 79.4              & 0.42                    & 63.7             & 1.15                   & 10            & 2.35                & 29.6              & 4.16                    \\ \hline
ELMo 150                   & 81.4           & 1.34                 & 70.8          & 2.17                & 76.6           & 0.55                 & 55.2          & 1.1                 & 79                & 0.87                    & 63               & 1.54                   & 5.8           & 1.1                 & 21.8              & 1.79                    \\ \hline
ELMo ft 150                & 83             & 1                    & 74            & 2.12                & 72.2           & 0.45                 & 47            & 0.71                & 77.6              & 0.42                    & 60.5             & 0.94                   & 12.8          & 1.3                 & 36.4              & 2.61                    \\ \hline
ELMo + BiLSTM 256          & 81.6           & 1.34                 & 72.2          & 2.17                & 72.4           & 2.3                  & 47.6          & 4.34                & 77                & 1.32                    & 59.9             & 2.51                   & 11.2          & 3.35                & 34                & 6.04                    \\ \hline
ELMo + BiLSTM 150          & 81.4           & 0.89                 & 71.4          & 1.82                & 72.2           & 1.3                  & 47.2          & 2.77                & 76.8              & 0.57                    & 59.3             & 1.15                   & 11.4          & 2.3                 & 33.8              & 4.97                    \\ \hline
ELMo ft + BiLSTM 150       & 82             & 1.22                 & 72.4          & 2.07                & 72             & 1.58                 & 46.6          & 2.61                & 77                & 1.37                    & 59.5             & 2.29                   & 12.2          & 0.84                & 35.6              & 2.3                     \\ \hline
DistilBERT 256             & 83.8           & 0.84                 & 75            & 1.87                & 76.8           & 1.1                  & 56            & 2.35                & 80.3              & 0.27                    & 65.5             & 1                      & 8.2           & 2.28                & 25.2              & 4.15                    \\ \hline
DistilBERT 150             & 80.4           & 0.55                 & 70.2          & 0.45                & 75.4           & 0.89                 & 53            & 2                   & 77.9              & 0.42                    & 61.6             & 1.14                   & 6.2           & 1.64                & 24.8              & 2.68                    \\ \hline
DistilBERT ft 256          & 83.2           & 0.84                 & 74.6          & 1.67                & 76.4           & 0.89                 & 55.4          & 1.82                & 79.8              & 0.76                    & 65               & 1.5                    & 8.2           & 1.3                 & 25.6              & 2.19                    \\ \hline
DistilBER ft 150           & 80             & 1.41                 & 69            & 3.24                & 71             & 2.55                 & 45.6          & 4.34                & 75.5              & 1.17                    & 57.3             & 2.14                   & 11            & 4.06                & 33.8              & 7.95                    \\ \hline
DistilBER BiLSTM 256       & 84.2           & 1.79                 & 76.4          & 3.05                & 78.4           & 2.61                 & 59            & 5.2                 & 81.3              & 2.11                    & 67.7             & 3.88                   & 7             & 1.87                & 23                & 5.39                    \\ \hline
DistilBERT BiLSTM 150      & 81.6           & 0.89                 & 71.8          & 1.3                 & 72.6           & 0.89                 & 48.2          & 2.17                & 77.1              & 0.82                    & 60               & 1.62                   & 11            & 0.71                & 32.8              & 2.39                    \\ \hline
DistilBERT ft + BiLSTM 256 & 84             & 1.58                 & 76.2          & 2.59                & 77.2           & 1.48                 & 57.4          & 2.19                & 80.6              & 1.52                    & 66.8             & 2.33                   & 8             & 0.71                & 24.8              & 1.3                     \\ \hline
DistilBERT ft + BiLSTM 150 & 81.8           & 0.84                 & 71.8          & 1.3                 & 72.2           & 1.48                 & 47.6          & 2.88                & 77                & 0.94                    & 59.7             & 1.96                   & 11.8          & 1.92                & 33.6              & 3.36                    \\ \hline
BERT 150         & 80.8           & 0.45                 & 70            & 0                   & 69             & 2.65                 & 41            & 4.85                & 74.9               & 1.29                    & 55.5             & 2.42                   & 14.8          & 3.42                & 41.4              & 6.66                    \\ \hline
BERT ft 150         & 81             & 0                    & 70.8          & 0.45                & 70.6           & 0.55                 & 44.8          & 0.84                & 75.8               & 0.27                    & 57.8             & 0.27                   & 12.8          & 1.1                 & 36.8              & 1.64                    \\ \hline
BERT + BiLSTM 150   & 81.8           & 1.1                  & 72            & 2.55                & 72.6           & 3.05                 & 48            & 5.79                & 77.2               & 1.92                    & 60               & 3.84                   & 11            & 2.92                & 33.6              & 6.58                    \\ \hline
BERT ft + BiLSTM 150 & 81.8           & 1.1                  & 72            & 2.55                & 72.6           & 3.05                 & 48            & 5.79                & 77.2               & 1.92                    & 60               & 3.84                   & 11            & 2.92                & 33.6              & 6.58                    \\ \hline
\end{tabular}
\end{adjustbox}
\label{table:det-pc}
\end{table}

\begin{table}[H]
\centering
\caption[Sentiment - detailed results.]{SA results including standard deviations. Macc: Accuracy on MR. Cacc: Accuracy on CR. Acc avg: Average of null and cross-context accuracies. Adrop: ``Drop'' score between null and cross-context accuracies.}
\vspace{1em}
\begin{adjustbox}{width=1\textwidth}
\begin{tabular}{|l|l|l|l|l|l|l|l|l|l|l|l|l|l|l|l|l|}
\hline
\textbf{model}            & \textbf{Ntest} & \textbf{Ntest stdev} & \textbf{Nacc} & \textbf{Nacc} & \textbf{Ctest} & \textbf{Ctest stdev} & \textbf{Cacc} & \textbf{Cacc stdev} & \textbf{Test avg} & \textbf{Test avg stdev} & \textbf{Acc avg} & \textbf{Acc avg stdev} & \textbf{Drop} & \textbf{Drop stdev} & \textbf{Adrop} & \textbf{Adrop stdev} \\ \hline
ELMo 60                   & 78               & 0.71             & 78            & 0.71                & 63.6             & 1.14                   & 64            & 0.71                & 70.8                & 0.76                      & 71               & 1                      & 18.4           & 1.14               & 18             & 0.71               \\ \hline
ELMo ft 60                & 76.2             & 0.45             & 76.2          & 0.45                & 69               & 1.58                   & 69.2          & 1.48                & 72.6                & 0.89                      & 72.7             & 1                      & 9.6            & 1.95               & 9.2            & 1.79               \\ \hline
ELMo + BiLSTM 60          & 79               & 0.71             & 79            & 0.71                & 67               & 2.65                   & 67            & 2.65                & 73                  & 1.5                       & 73               & 2                      & 15.2           & 3.27               & 15.2           & 3.27               \\ \hline
ELMo ft + BiLSTM 60       & 78.2             & 0.84             & 67.4          & 0.84                & 67.4             & 1.34                   & 67.6          & 1.67                & 72.8                & 1.04                      & 72.9             & 1                      & 13.8           & 0.84               & 13.4           & 1.52               \\ \hline
DistilBERT 60             & 79               & 0.71             & 79            & 0.71                & 66.8             & 3.96                   & 67            & 3.54                & 72.9                & 2.27                      & 73               & 2                      & 15.4           & 4.51               & 15.2           & 4.09               \\ \hline
DistlBERT ft 60           & 79               & 0.71             & 79            & 0.71                & 68               & 1.22                   & 68            & 1.22                & 73.5                & 0.5                       & 73.5             & 1                      & 14.2           & 2.17               & 14.2           & 2.17               \\ \hline
DistilBERT + BiLSTM 60    & 80               & 1                & 80            & 1                   & 70.2             & 2.28                   & 70.2          & 2.28                & 75.1                & 1.64                      & 75.1             & 2                      & 12.4           & 1.82               & 12.4           & 1.82               \\ \hline
DistilBERT ft + BiLSTM 60 & 80               & 1                & 80            & 1                   & 70.2             & 2.28                   & 70.2          & 2.28                & 75.1                & 1.64                      & 75.1             & 2                      & 12.4           & 1.82               & 12.4           & 1.82               \\ \hline
BERT 60             & 80.6           & 0.55                 & 80.6          & 0.55                & 75             & 1.58                 & 75            & 1.58                & 77.8               & 1.04                    & 77.8             & 1.04                   & 7             & 1.58                & 7              & 1.58                 \\ \hline
BERT ft 60          & 80.6           & 0.55                 & 80.6          & 0.55                & 75             & 1.58                 & 75            & 1.58                & 77.8               & 1.04                    & 77.8             & 1.04                   & 7             & 1.58                & 7              & 1.58                 \\ \hline
BERT + BiLSTM 60    & 82             & 1                    & 82            & 1                   & 71.8           & 3.56                 & 71.8          & 3.56                & 76.9               & 2.13                    & 76.9             & 2.13                   & 12.6          & 3.91                & 12.6           & 3.91                 \\ \hline
BERT ft + BiLSTM 60 & 82.6           & 0.55                 & 82.6          & 0.55                & 75.6           & 1.34                 & 75.8          & 1.64                & 79.1               & 0.89                    & 79.2             & 1.04                   & 8.6           & 1.52                & 8.4            & 1.82                 \\ \hline
\end{tabular}
\end{adjustbox}
\label{table:det-sa}
\end{table}

\chapter{Hyper-parameter tuning results}
\label{chapter:hyp-tune}

\begin{table}[H]
\centering
\caption[Hyper-parameter tuning results.]{Hyper-parameter tuning results.}
\vspace{1em}
\begin{adjustbox}{width=1\textwidth}
\begin{tabular}{|l|l|l|l|l|l|l|l|l|l|l|}
\hline
\textbf{task} & \textbf{model}             & \textbf{opt best train F-score} & \textbf{opt best dev F-score} & \textbf{opt best epoch no} & \textbf{opt time (hrs:mins)} & \textbf{drop} & \textbf{l2} & \textbf{lrate} & \textbf{lr decay} & \textbf{relu} \\ \hline
PC            & ELMo 256                   & 90                             & 87                           & 9                          & 6:26                         & 0.25          & 0           & 1e-3       & 0.5               & TRUE          \\ \hline
PC            & ELMo 150                   & 81                             & 87                           & 5                          & 4:11                         & 0             & 0.01        & 1e-3       & 0.5               & TRUE          \\ \hline
PC            & ELMo ft 150                & 91                             & 87                           & 9                          & 9:21                         & 0             & 0           & 1e-3       & 0                 & TRUE          \\ \hline
PC            & ElMo + BiLSTM 256          & 84                             & 84                           & 7                          & 8:37                         & 0             & 0           & 5e-5       & 0.5               & TRUE          \\ \hline
PC            & ElMo + BiLSTM 150          & 89                             & 86                           & 9                          & 4:11                         & 0.5           & 0.01        & 1e-3       & 0                 & FALSE         \\ \hline
PC            & ElMo ft + BiLSTM 150       & 86                             & 87                           & 5                          & 9:03                         & 0             & 0.01        & 1e-3       & 0.5               & FALSE         \\ \hline
PC            & DistilBERT 256             & 87                             & 87                           & 8                          & 1:49                         & 0             & 0           & 1e-3       & 0                 & TRUE          \\ \hline
PC            & DistilBERT 150             & 83                             & 85                           & 2                          & 1:05                         & 0             & 0           & 1e-3       & 0                 & TRUE          \\ \hline
PC            & DistilBERT ft 256          & 87                             & 87                           & 8                          & 1:52                         & 0.25          & 0           & 1e-3       & 0.5               & TRUE          \\ \hline
PC            & DistilBERT ft 150          & 78                             & 85                           & 5                          & 1:06                         & 0.25          & 0           & 1e-1       & 0                 & FALSE         \\ \hline
PC            & DistilBERT + BiLSTM 256    & 88                             & 84                           & 7                          & 2:24                         & 0             & 0.01        & 1e-3       & 0                 & TRUE          \\ \hline
PC            & DistilBERT + BiLSTM 150    & 87                             & 85                           & 7                          & 1:27                         & 0.5           & 0           & 5e-5       & 0                 & FALSE         \\ \hline
PC            & DistilBERT ft + BiLSTM 256 & 88                             & 86                           & 6                          & 2:27                         & 0             & 0.01        & 1e-3          & 0                 & FALSE         \\ \hline
PC            & DistilBERT ft + BiLSTM 150 & 87                             & 85                           & 7                          & 1:27                         & 0.5           & 0           & 5e-5       & 0.5               & FALSE         \\ \hline
PC & BERT 150             & 85 & 86 & 9 & 4:31 & 0.5  & 0    & 1e-3 & 0   & FALSE \\ \hline
PC & BERT ft 150          & 87 & 87 & 9 & 4:28 & 0.25 & 0    & 1e-3 & 0   & TRUE  \\ \hline
PC & BERT + BiLSTM 150    & 88 & 85 & 6 & 5:07 & 0.25 & 0.01 & 1e-3 & 0.5 & TRUE  \\ \hline
PC & BERT ft + BiLSTM 150 & 88 & 85 & 6 & 5:15 & 0.25 & 0.01 & 1e-3 & 0.5 & TRUE  \\ \hline
SA            & ELMo 60                    & 84                             & 80                           & 6                          & 3:56                         & 0             & 0           & 1e-3          & 0.5               & TRUE          \\ \hline
SA            & ELMo ft 60                 & 78                          & 78                        & 6                          & 6:44                         & 0.5           & 0           & 1e-3          & 0                 & FALSE         \\ \hline
SA            & ELMo + BiLSTM 60           & 83                             & 83                           & 2                          & 3:06                         & 0             & 0           & 1e-3          & 0                 & FALSE         \\ \hline
SA            & ELMo ft + BiLSTM 60        & 84                          & 82                        & 2                          & 9:49                         & 0             & 0           & 1e-3          & 0                 & TRUE          \\ \hline
SA            & DistilBERT 60              & 82                             & 79                           & 5                          & 1:10                         & 0             & 0           & 1e-3          & 0.5               & TRUE          \\ \hline
SA            & DistilBERT ft 60           & 82                             & 79                           & 5                          & 1:32                         & 0.25          & 0           & 1e-3          & 0.5               & TRUE          \\ \hline
SA            & DistilBERT + BiLSTM 60     & 99                             & 84                           & 8                          & 1:34                         & 0             & 0           & 1e-3          & 0                 & FALSE         \\ \hline
SA            & DistilBERT ft + BiLSTM 60  & 99                             & 84                           & 8                          & 1:08                         & 0             & 0           & 1e-3          & 0                 & FALSE         \\ \hline
SA & BERT 60             & 82 & 82 & 8 & 2:16 & 0.5 & 0 & 5e-5 & 0   & TRUE  \\ \hline
SA & BERT ft 60          & 82 & 82 & 8 & 4:36 & 0.5 & 0 & 5e-5 & 0   & TRUE  \\ \hline
SA & BERT + BiLSTM 60    & 88 & 84 & 2 & 2:33 & 0   & 0 & 1e-3 & 0   & TRUE  \\ \hline
SA & BERT ft + BiLSTM 60 & 99 & 85 & 8 & 5:10 & 0.5 & 0 & 1e-3 & 0.5 & FALSE \\ \hline
\end{tabular}
\end{adjustbox}
\label{table:hyp-tune}
\end{table}

\end{document}